\pdfoutput=1
\documentclass[twoside,11pt]{article}

\usepackage[preprint]{jmlr2e}
\usepackage{xurl}
\usepackage{amsmath}
\usepackage{booktabs}
\usepackage{array}
\usepackage{lastpage} 
\usepackage{float}

\newcommand{\R}{\mathbb{R}}
\newcommand{\psims}{Physics-SIMS-TS}
\newcommand{\simsts}{SIMS-TS}

\jmlrheading{1}{2026}{1-\pageref{LastPage}}{9/26}{--}{00-0000}{Temesgen Mikael Abraha and Yves Lucet}
\ShortHeadings{Monotone-Constrained Diffusion Forecasting}{Abraha and Lucet}
\firstpageno{1}

\begin{document}

\title{Monotone-Constrained Diffusion Models for\\
Long-Horizon Production Forecasting}

\author{\name Temesgen Mikael Abraha \email temesgen.abraha@ubc.ca \\
  \addr Department of Computer Science, I.\,K.\,Barber Faculty of Science\\
  University of British Columbia\\
  Kelowna, BC V1V~1V7, Canada
  \AND
  \name Yves Lucet \email yves.lucet@ubc.ca \\
  \addr Department of Computer Science, I.\,K.\,Barber Faculty of Science\\
  University of British Columbia\\
  Kelowna, BC V1V~1V7, Canada}

\editor{}

\maketitle

\begin{abstract}%
Forecasting a long horizon from only the first observations of a sequence is ill-posed: many trajectories are consistent with the same short history. We study this problem in oil and gas production forecasting, where forecasts made after roughly the first fifth of a well's producing life drive development and abandonment decisions, and where a usable forecast must describe a monotone decline. We present \psims{}, a conditional diffusion forecaster that combines negative guidance against synthetic artifacts, decline-curve constraints and an isotonic projection applied during sampling, spatial training augmentation, and an ensembled stochastic sampler yielding a full predictive distribution. Across three jurisdictions and more than 35{,}000 wells, under a shared-space, validation-frozen protocol, \psims{} is the most accurate diffusion forecaster in the comparison and is competitive with, but not superior to, ensembled transformer forecasters. Its forecasts are monotone by construction at a cost of at most 0.5\% in mean squared error, and its trajectory ensemble yields calibrated intervals after one dispersion factor is fitted per jurisdiction. On six standard benchmarks a reversible-instance-normalization variant of the backbone is the leading diffusion baseline. We also quantify four protocol choices on which the measured ranking depends. Code and evaluation artifacts are released.
\end{abstract}

\begin{keywords}
  diffusion models, time series forecasting, probabilistic forecasting, shape-constrained prediction, production forecasting
\end{keywords}

\section{Introduction}
\label{sec:introduction}

Many forecasting problems require predicting a long horizon from a short history. When only the first fraction of a sequence is observed, the task is ill-posed, because many full trajectories are consistent with the same early observations. A forecaster must then choose among them using whatever structure it can exploit, whether learned from related sequences or supplied as prior knowledge. The difficulty is most acute in safety-critical industrial settings, where decisions are made early and the cost of an inaccurate forecast is large.

Oil and gas production forecasting is one such setting, and it motivates this work. The classical approach fits the Arps decline-curve equations~\citep{arps1945analysis} to a well's history, but reliable parameter estimates traditionally require most of a well's producing life, whereas completion, development, and reserve decisions are made from the first years of production. In the jurisdictions studied here, that is roughly the first fifth of a well's eventual producing life, about two years of data. A forecast also determines when a well reaches its economic limit and should be scheduled for abandonment and reclamation. Wells without a reliable forecast risk being left idle and eventually orphaned, with no solvent party responsible for cleanup. British Columbia had 987 designated orphan sites in September 2025, about a quarter of them fully reclaimed~\citep{bcer2025erikson}, the Orphan Well Association's inventory held 4{,}200 wells awaiting decommissioning at the end of March 2026~\citep{owa2026annual}, and the United States had at least 123{,}000 documented orphaned wells as of April 2022~\citep{kang2022orphaned}. Accurate early forecasts therefore carry direct economic and environmental consequences.

The underlying methodological question is general. Given many complete sequences for training and only a short prefix of a new sequence at test time, can a generative model constrained by domain knowledge during sampling forecast the remaining horizon more accurately than data-driven methods alone? We study this question with diffusion models. Denoising diffusion probabilistic models~\citep{ho2020denoising} generate samples by reversing a gradual noising process and have produced strong results in time-series generation, imputation, and forecasting~\citep{diffusion-ts2024, tsdiff2023, tashiro2021csdi}. A known failure mode is that training a generative model on its own outputs causes progressive degradation, termed model autophagy disorder~\citep{alemohammad2024mad} or model collapse~\citep{shumailov2024ai}. The Self-Improving diffusion Models with Synthetic data (SIMS) framework~\citep{sims2024} avoids this by using synthetic data as \emph{negative} guidance, steering generation away from synthetic artifacts rather than toward them. SIMS was developed for image generation and, to our knowledge, has not been adapted to time series, nor has negative guidance been combined with physics constraints for forecasting.

We adapt SIMS to conditional time-series forecasting and add physics guidance derived from decline-curve dynamics. The resulting method, \psims{}, observes the first 20\% of a sequence and predicts the remaining 80\%. We make the following contributions.

\begin{enumerate}
\item We present \psims{}, a conditional diffusion forecaster that is the most accurate member of its family on this task. Against CSDI, Diffusion-TS, TSDiff, and TimeGrad, at matched training budgets and one trained model each, it attains the lower error in all twelve baseline--jurisdiction comparisons, on both metrics, and on 60 to 82\% of individual wells. Under the full protocol, it is competitive with, but not superior to, ensembled transformer forecasters.
\item We make monotone decline a guarantee rather than a tendency, through decline-curve constraints during sampling followed by an isotonic projection, and measure its cost at no more than 0.5\% in mean squared error. The guarantee, not accuracy, is what the constraints provide: the projection is exportable to any forecaster at comparable cost, and the sampling-time guidance is accuracy-neutral.
\item We show that the trajectory ensemble supports interval estimation. Raw intervals are overconfident, but a single dispersion factor per jurisdiction, fitted on one half of the test wells and evaluated on the other, restores near-nominal coverage at every level tested and yields a joint distribution over whole curves rather than pointwise bands.
\item We identify four protocol choices on which the measured method ranking depends: the normalization space in which error is computed, the training-epoch budget, whether baselines receive the same pretrain-and-fine-tune pipeline, and whether the matched budget extends to the transfer and leave-one-domain-out studies. Each is quantified by comparing the two settings on identical data, and in each case the less controlled setting favours the diffusion model.
\end{enumerate}

A preliminary version of this work appeared at the Canadian AI Conference 2026~\citep{abraha2026physics}. This article differs in scope and in evaluation methodology. In scope, it adds the Alberta and Pennsylvania jurisdictions, the transfer and leave-one-domain-out studies, the standard-benchmark and multi-seed studies, and the probabilistic evaluation. In methodology, the conference version, like much prior work in this application area, evaluated each method under its own normalization, whereas this article evaluates all methods in a single shared space, selects all configurations on validation data under a frozen protocol, and grants the strongest baselines matched ensemble budgets. Section~\ref{sec:protocol-effects} shows that these choices materially affect cross-method comparisons, so the absolute values and margins reported here supersede those of the conference version~\citep{abraha2026physics}, whose qualitative conclusion, that constrained diffusion forecasting is competitive in this regime while guaranteeing physically admissible forecasts, is supported under the stricter protocol.

The remainder of the paper is organized as follows. Section~\ref{sec:related_work} reviews related work, Section~\ref{sec:method} presents \psims{}, Section~\ref{sec:setup} describes the data, baselines, and protocol, Section~\ref{sec:results} reports results, and Sections~\ref{sec:limitations} and~\ref{sec:conclusion} discuss limitations and conclude.

\section{Related Work}
\label{sec:related_work}

\paragraph{Decline-curve analysis and physics-informed forecasting.}
The Arps empirical decline equations~\citep{arps1945analysis} remain the industry standard for production forecasting over the oil and gas well lifecycle~\citep{tadjer2021machine}. Extensions such as the stretched-exponential model~\citep{valko2009stretched} and the Duong model~\citep{duong2011approach} address transient flow in unconventional reservoirs, where the several flow regimes involved can make the Arps model a poor fit~\citep{tadjer2021machine}. Physics-informed neural networks have been applied to reservoir problems with sparse data~\citep{han2023physics}, and flow theory has been embedded in prediction loss functions~\citep{bi2024physics}. We instead inject decline-curve physics into the sampling process of a diffusion model, which keeps the generative prior intact while enforcing physical plausibility at inference.

\paragraph{Diffusion models for time series.}
Diffusion models achieve strong results in time-series generation and forecasting. Diffusion-TS~\citep{diffusion-ts2024} uses encoder-decoder transformers with disentangled seasonal-trend decomposition; TSDiff~\citep{tsdiff2023} adds self-guidance for unconditionally trained models; CSDI~\citep{tashiro2021csdi} conditions on observed values through two-dimensional attention; and TimeGrad~\citep{timegrad2021} couples autoregressive prediction with denoising diffusion. \citet{shen2023timediff} show that a conditional diffusion forecaster benefits from an autoregressive initial estimate of the horizon supplied as conditioning, and \citet{meng2022sdedit} show that the reverse process can start from a noised guide rather than from pure noise; Section~\ref{sec:psims-config} combines the two. These architectures form the baselines against which we compare, and the conditional formulation of CSDI is closest to our backbone.

\paragraph{Transformer and foundation-model forecasters.}
Transformer forecasters define the current accuracy frontier on standard long-horizon benchmarks. PatchTST~\citep{patchtst2023} segments each series into patches and forecasts each channel independently, and iTransformer~\citep{itransformer2024} attends across variates rather than across time. Time-series foundation models such as Chronos~\citep{chronos2024} and its successor Chronos-2~\citep{chronos2_2025} are pretrained for zero-shot forecasting. We include all four as baselines to situate a diffusion-based method against the strongest current alternatives.

\paragraph{Data augmentation and spatial interpolation.}
\citet{wen2021time} provide a taxonomy of time-series augmentation, and \citet{iwana2021empirical} evaluate twelve techniques across 128 datasets. Generative augmentation includes TimeGAN~\citep{timegan2019} and TimeVAE~\citep{timevae2021}; the TSGBench study~\citep{ang2024tsgbench} finds that variational approaches rank well on distance metrics while TimeGAN underperforms its reputation. For spatially distributed measurements, inverse distance weighting (IDW)~\citep{shepard1968two} estimates unsampled values as distance-weighted averages of neighbors. Augmentation through generative models for production data remains largely unexplored. Recent surveys of machine learning for production forecasting concentrate on regression and physics-informed approaches rather than generative augmentation~\citep{fan2025review}.

\paragraph{Model collapse and SIMS.}
Model autophagy disorder~\citep{alemohammad2024mad}, also called model collapse~\citep{shumailov2024ai}, describes the degradation that occurs when generative models train on their own outputs. SIMS~\citep{sims2024} addresses this through negative guidance and maintains stable performance across 100 self-consuming generations on a two-dimensional Gaussian benchmark. We borrow the mechanism rather than the multi-generation setting: our auxiliary model is trained on a single generation of synthetic sequences, so negative guidance acts here as a correction away from the base model's own sampling artifacts, not as a demonstrated remedy for collapse across generations. Whether the self-consuming stability reported for images carries to time series is outside the scope of this article; Section~\ref{sec:ablation} reports what the mechanism contributes at our operating point.

\section{Method}
\label{sec:method}

\subsection{\simsts{}: Negative Guidance for Time Series}
\label{sec:simsts}

We model a normalized sequence $\mathbf{x} \in \R^{100}$ representing a well's production history. Each well's complete producing history is linearly interpolated onto a fixed grid of 100 points, so the grid index measures the fraction of producing life; months with no reported production are dropped before interpolation. Each curve is then min-max normalized \emph{per curve} to $[0,1]$, the same normalization used by every baseline, so that all methods in this article are trained and evaluated in a single shared space (Section~\ref{sec:protocol-effects} motivates this choice). We observe the first 20 points $\mathbf{c} = \mathbf{x}_{1:20}$ and predict the remaining 80 points $\mathbf{y} = \mathbf{x}_{21:100}$: the first 20\% of a well's producing life, typically around two years of data, determines the forecast of the remaining 80\%.

We use a conditional denoising diffusion probabilistic model whose forward process adds Gaussian noise according to
\begin{equation}
\label{eq:forward}
q(\mathbf{x}_t \mid \mathbf{x}_0) = \mathcal{N}\!\left(\mathbf{x}_t;\, \sqrt{\bar{\alpha}_t}\, \mathbf{x}_0,\; (1 - \bar{\alpha}_t)\mathbf{I}\right),
\end{equation}
with cumulative noise schedule $\bar{\alpha}_t = \prod_{s=1}^{t}(1 - \beta_s)$. The denoising network $\epsilon_\theta(\mathbf{x}_t, t, \mathbf{c})$ is a transformer with adaptive layer normalization that injects the diffusion timestep $t$ and the conditioning signal $\mathbf{c}$ through additive embeddings. Training minimizes the score-matching objective
\begin{equation}
\label{eq:score}
\mathcal{L} = \mathbb{E}_{t, \mathbf{x}_0, \epsilon}\!\left[\,\|\epsilon - \epsilon_\theta(\mathbf{x}_t, t, \mathbf{c})\|^2\,\right].
\end{equation}

The key idea of SIMS~\citep{sims2024} is that synthetic data should provide negative rather than positive guidance. We train a base model $\epsilon_{\theta_r}$ on real sequences, generate synthetic sequences from it under the same conditioning prefixes, and train an auxiliary model $\epsilon_{\theta_s}$ on those synthetic samples under the identical recipe. At inference, the two predictions are combined through
\begin{equation}
\label{eq:sims}
\tilde{\epsilon} = (1 + \omega)\, \epsilon_{\theta_r}(\mathbf{x}_t, t, \mathbf{c}) - \omega\, \epsilon_{\theta_s}(\mathbf{x}_t, t, \mathbf{c}),
\end{equation}
where $\omega \geq 0$ controls the guidance strength. Equation~\eqref{eq:sims} amplifies the direction pointing from the synthetic distribution toward the real one, which counteracts the drift that causes model collapse.

\subsection{\psims{}: Integrating Decline-Curve Constraints}
\label{sec:physics}

Production sequences approximately follow Arps decline dynamics and decrease monotonically. The Arps hyperbolic model describes the rate $q(k)$ at production step $k$ through
\begin{equation}
\label{eq:arps}
q(k) = \frac{q_i}{(1 + b\, D_i\, k)^{1/b}},
\end{equation}
where $q_i$ is the initial rate, $D_i$ the initial decline rate, and $b$ the decline exponent; $k$ indexes production steps, to distinguish it from the diffusion timestep $t$.

We encode these properties through three penalty terms evaluated during the reverse process, all computed directly on the estimated clean sequence $\hat{\mathbf{x}}_0$ in the normalized space, with the step index rescaled to the unit interval. The monotonicity and non-negativity terms are exact in any monotonically transformed space. The decline term is a \emph{shape prior}, a soft penalty pulling trajectories toward the hyperbolic decline,
\begin{equation}
\label{eq:arps-ode}
\mathcal{L}_{\text{shape}} = \frac{1}{K} \sum_{k=1}^{K} \left| \frac{d\hat{q}_k}{d\kappa} + \tilde{D}(\kappa_k)\, \hat{q}_k \right|^2,
\qquad
\tilde{D}(\kappa) = \frac{\tilde{D}_i}{1 + b\, \tilde{D}_i\, \kappa},
\end{equation}
where $\kappa \in [0,1]$ is the rescaled step index, $K$ is the sequence length, and derivatives use central finite differences (one-sided at the endpoints). The prior parameters $(q_i, \tilde{D}_i, b)$ are estimated at the formation level by a bounded least-squares fit of \eqref{eq:arps} to the mean training target curve of the formation (for Pennsylvania, $\tilde{D}_i = 2.24$, $b = 0.52$). Because only the first 20\% of a well is observed at test time, per-well fitting on the conditioning prefix would be unreliable; the formation-level fit supplies representative decline dynamics without touching any test well. The monotonicity term penalizes any increase,
\begin{equation}
\label{eq:mono}
\mathcal{L}_{\text{mono}} = \frac{1}{K-1} \sum_{k=1}^{K-1} \max\!\left(0,\, \hat{q}_{k+1} - \hat{q}_k\right)^2,
\end{equation}
and the boundary term enforces non-negativity, $\mathcal{L}_{\text{bound}} = \frac{1}{K} \sum_{k=1}^{K} \max(0, -\hat{q}_k)^2$.

At each denoising step we compute the SIMS-guided prediction $\tilde{\epsilon}$ from \eqref{eq:sims}, estimate $\hat{\mathbf{x}}_0 = (\mathbf{x}_t - \sqrt{1-\bar{\alpha}_t}\,\tilde{\epsilon}) / \sqrt{\bar{\alpha}_t}$, and apply gradient guidance
\begin{equation}
\label{eq:guidance}
\tilde{\epsilon} \leftarrow \tilde{\epsilon} + \eta(t)\, \mathbf{g}, \qquad
\mathbf{g} = \frac{\nabla_{\hat{\mathbf{x}}_0} \mathcal{L}_{\text{physics}}}{\max\!\left(1, \|\nabla_{\hat{\mathbf{x}}_0} \mathcal{L}_{\text{physics}}\|\right)},
\end{equation}
with $\mathcal{L}_{\text{physics}} = \lambda_1 \mathcal{L}_{\text{shape}} + \lambda_2 \mathcal{L}_{\text{mono}} + \lambda_3 \mathcal{L}_{\text{bound}}$ and $(\lambda_1, \lambda_2, \lambda_3) = (1, 2, 1)$. Because $\hat{\mathbf{x}}_0$ decreases by $\sqrt{(1-\bar{\alpha}_t)/\bar{\alpha}_t}$ per unit added to $\tilde{\epsilon}$, \eqref{eq:guidance} is a descent step on $\mathcal{L}_{\text{physics}}$ in $\hat{\mathbf{x}}_0$-space; the timestep-dependent factor is absorbed into the tuned strength $\eta_0$, and the norm clip bounds any single correction. The guidance strength follows the decaying schedule
\begin{equation}
\label{eq:adaptive-eta}
\eta(t) = \eta_0 \left(\frac{t}{T}\right)^{1/2},
\end{equation}
which is strongest early in sampling, when a coarse shape correction selects among the many trajectories consistent with the conditioning, and decays as samples become cleaner, where strong gradients can destabilize nearly denoised sequences. The hard guarantee is supplied at the end of sampling: every generated sequence is projected onto the set of monotonically decreasing sequences by isotonic regression through the pool-adjacent-violators (PAV) algorithm~\citep{isotonic1972}. Physics enters at inference only; training uses the standard objective of \eqref{eq:score}.

\subsection{Training and Inference Configuration}
\label{sec:psims-config}

Three standard components complete the method. First, the training set is enlarged five-fold with the IDW spatial augmentation of Section~\ref{sec:aug}, using the same settings as the IDW+RF baseline, which pairs IDW with a random forest (RF) regressor (power 3, up to 8 neighbors): synthetic training wells are spatial blends of neighboring wells' curves, generated once before training. Second, training uses an exponential moving average of the network weights (decay 0.999) and minimum signal-to-noise-ratio (min-SNR) weighting of the diffusion loss~\citep{hang2023minsnr} with $\gamma = 5$. Third, \psims{} is a probabilistic forecaster: its predictive distribution is represented by $K = 20$ stochastic denoising diffusion implicit model (DDIM) samples~\citep{song2020ddim} at noise level $\eta_{\mathrm{DDIM}} = 0.5$, drawn from each of $R = 5$ independently trained models~\citep{lakshminarayanan2017deep}, and its point forecast is the pointwise mean of the resulting 100 trajectories, each already projected by the PAV step at the end of sampling, followed by a cumulative minimum. Sampling starts partway into the reverse trajectory from a noised prior rather than from pure noise~\citep{meng2022sdedit}: the prior is the formation's mean training target curve scaled to the well's last observed value, noised to the level corresponding to 70\% of the trajectory, in the spirit of the autoregressive initialization of \citet{shen2023timediff}, which there enters as conditioning rather than as the starting point of sampling. Section~\ref{sec:ablation} quantifies the contribution of each component, and Section~\ref{sec:protocol} states how all values were selected.

\subsection{Backbone Variant for Non-Stationary Series}
\label{sec:revin}

The production case study uses sequences that share a common decline shape, so the backbone of \eqref{eq:forward}--\eqref{eq:score} trains directly on normalized curves. Standard forecasting benchmarks include strongly non-stationary series whose level and scale shift between the lookback window and the forecast horizon, a regime in which diffusion forecasters have historically been weak. For these benchmarks, we apply reversible instance normalization (RevIN)~\citep{revin2022} to the input window: RevIN removes the per-window mean and scale before the model sees the input and restores them at the output, which neutralizes the level shift. We further replace the deterministic sampler with a five-sample stochastic DDIM ensemble at noise level $\eta_{\mathrm{DDIM}} = 0.5$ and take the median, which reduces trajectory-level variance roughly independently of RevIN. The physics-guided sampler of Section~\ref{sec:physics} is specific to decline curves and is not applied to the standard benchmarks; the RevIN variant therefore isolates the quality of the diffusion backbone itself.

\section{Experimental Setup}
\label{sec:setup}

\subsection{Datasets}
\label{sec:datasets}

\paragraph{Production case study.}
The primary data come from 16{,}216 gas wells in British Columbia, Canada, obtained from the British Columbia Energy Regulator, the Crown corporation that regulates oil and gas activity in the province~\citep{bcgov2026bcer}, spanning the Montney, Doig, Charlie Lake, Baldonnel, Debolt, Golata, and Wilrich formations. Only wells with at least 24 producing months are retained, and each well's complete producing history is interpolated onto the fixed 100-point grid of Section~\ref{sec:simsts}; one grid step therefore corresponds to $1/100$ of the individual well's producing life, not to a fixed calendar duration. We organize wells into formation-based clusters ranging from 51 to 16{,}216 wells, which lets us measure behavior across three orders of magnitude in dataset size. To test cross-jurisdiction generalization, we add two further unconventional plays: Alberta Montney (6{,}747 wells) and Pennsylvania (PA) Marcellus (12{,}360 wells, built by joining the Pennsylvania Department of Environmental Protection production reports~\citep{padep2026} to the Pennsylvania Spatial Data Access spatial header~\citep{pasda2026}). The three jurisdictions together comprise more than 35{,}000 wells.  

Median producing histories are 136 months in British Columbia, 126 in Pennsylvania, and 93 in Alberta, so the 20-point conditioning prefix typically corresponds to about 27, 25, and 19 producing months, respectively. In every jurisdiction, wells are split 80/20 into training and test sets by well identifier under a fixed ordering and seed, so no well contributes to both sides, and all reported metrics are computed on the held-out test wells. Per-curve normalization uses each curve's own range and no cross-well statistics. Because the normalizing minimum of a declining curve typically lies in the forecast region (81 to 89\% of test wells, depending on jurisdiction), normalized errors share a mild dependence on each well's final rate that is common to every method compared; metrics in the data units appear alongside the normalized metrics in Section~\ref{sec:leaderboard}. On the normalized scale, a mean squared error (MSE) of 0.01 corresponds to a root mean squared error of 0.10, roughly 10\% of a well's production range.

\paragraph{Standard benchmarks.}
To measure the diffusion backbone outside the production domain we use six standard forecasting datasets: ETTh1, ETTh2, ETTm1, ETTm2, Exchange, and Weather. We follow the common protocol with lookback 96, horizons $H \in \{96, 192, 336, 720\}$, and channel-independent evaluation, giving 24 settings.

\subsection{Baselines}
\label{sec:baselines}

We compare \psims{} against ten baselines, grouped by family. The classical baseline is the Arps decline-curve model of \eqref{eq:arps}, fitted by bounded least squares to each well's observed prefix and extrapolated over the horizon (Section~\ref{sec:leaderboard}). The traditional baseline, IDW+RF, is IDW augmentation with a random forest regressor. The transformer baselines are PatchTST~\citep{patchtst2023} and iTransformer~\citep{itransformer2024}. The diffusion baselines are CSDI~\citep{tashiro2021csdi}, Diffusion-TS~\citep{diffusion-ts2024}, TSDiff~\citep{tsdiff2023}, and TimeGrad~\citep{timegrad2021}; \psims{} extends this family, and \simsts{}, which is \psims{} without the physics terms, appears throughout as its ablation rather than as an independent baseline. The foundation-model baselines are Chronos-2~\citep{chronos2_2025} and Chronos-Bolt, a patch-based, direct multi-step variant released with the Chronos code base~\citep{chronos2024, chronosbolt2024}, evaluated zero-shot. For the augmentation study of Section~\ref{sec:aug}, we additionally compare IDW against the generative augmenters TimeGAN~\citep{timegan2019} and TimeVAE~\citep{timevae2021}, holding the downstream regressor fixed so that the comparison isolates the augmentation method rather than regressor capacity. Random forests and gradient boosting are standard downstream regressors for production forecasting~\citep{fan2025review}, which is why we adopt them in that controlled comparison.

\subsection{Implementation and Training}
\label{sec:impl}

The denoising network is a transformer with hidden dimension 128, four layers, four attention heads, and sinusoidal timestep embeddings combined additively with a learned encoder for the observed 20-point prefix (1.26M parameters). The noise schedule uses 1{,}000 linearly spaced timesteps. Each model is trained with Adam (learning rate $10^{-4}$, cosine annealing, batch size 64, gradient clipping at norm 1.0) for at most 200 epochs, with 10\% of the augmented training pool held out for early stopping at patience 30; this block controls training length rather than providing a clean held-out estimate, and the deployed weights are the exponential moving average. The auxiliary model of \eqref{eq:sims} is trained on up to 8{,}000 sequences generated by the base model under the same recipe. For clusters with fewer than 2{,}000 wells, both models are initialized from the corresponding pretrained models and fine-tuned on cluster-specific data for 50 epochs at learning rate $10^{-5}$; the pretraining wells exclude every cluster's test wells, so no cluster result depends on a model that has seen its own test data. Diffusion sampling uses DDIM~\citep{song2020ddim} with 50 steps, and the physics loss weights are $\lambda_1 = 1.0$, $\lambda_2 = 2.0$, $\lambda_3 = 1.0$ throughout. On the standard benchmarks, \simsts{} is reported with the guidance strength fixed at $\omega = 0$ for every dataset and horizon; no per-setting hyperparameter selection is performed, so all benchmark models are compared under fixed configurations chosen in advance.

\subsection{Configuration and Selection Protocol}
\label{sec:protocol}

All configurable values of \psims{}, namely the guidance strengths $\omega$ and $\eta_0$, the per-model sample count $K$, the initialization depth, and the aggregation rule, were selected on a validation split of \emph{Pennsylvania training wells} (12.5\% of the training set, disjoint from all test wells) and then frozen. The frozen configuration ($\omega = 0.1$, $\eta_0 = 0.3$, $K = 20$, initialization depth 0.7, mean aggregation, $R = 5$ following the deep-ensemble convention of \citet{lakshminarayanan2017deep}) was fixed before any Alberta or British Columbia experiment was run and applied without change to those jurisdictions, to the formation clusters of Section~\ref{sec:clusters}, and to the transfer and leave-one-domain-out studies; the final models were retrained on each full training split under this configuration. The pipeline's component composition was developed on the Pennsylvania dataset, and Appendix~\ref{app:protocol} reports a confirmation on a second Pennsylvania train/test split drawn with a different seed. No baseline is tuned per dataset either: each baseline runs the size-adaptive configuration rule shipped with its implementation, which sets width, depth, batch size, and learning rate from dataset size, with early stopping on held-out training data and no per-dataset search.

\paragraph{Sampling budgets and point forecasts.}
Every probabilistic method is evaluated from multiple samples. The diffusion baselines (TimeGrad, CSDI, Diffusion-TS, TSDiff) each draw 10 stochastic samples per test well and report the per-timestep median, in all three jurisdictions, and their reported inference times include this ensembling. Chronos-Bolt and Chronos-2 output quantile forecasts, and we use their native median. \psims{} uses the $R \times K = 100$-trajectory predictive ensemble of Section~\ref{sec:psims-config}. Because multi-model ensembling is a generic device, Section~\ref{sec:leaderboard} additionally reports the strongest baselines under the same budget, five independently trained (or, for IDW+RF, independently seeded) instances per method with point forecasts averaged, and all conclusions in this article are stated against those matched-budget baselines; the remaining baselines are evaluated as single trained models and reported separately. The matched budget equalizes the number of instances and the training-epoch budget, not the size of the training set: \psims{} and IDW+RF train on the five-fold spatially augmented sets of Section~\ref{sec:aug} (64{,}780, 26{,}985, and 49{,}440 sequences on British Columbia, Alberta, and Pennsylvania), while the transformer and diffusion baselines train on the 12{,}972, 5{,}397, and 9{,}888 real training curves, because the augmentation is a component of the two spatial methods rather than of the protocol. That asymmetry favours \psims{}, and the transformers lead on mean squared error notwithstanding it.

\paragraph{Hardware.}
All experiments ran on two NVIDIA RTX 6000 Ada GPUs (48\,GB each); reported timings are single-GPU wall-clock times on this hardware.

\subsection{Evaluation Metrics}
\label{sec:metrics}

We report MSE as the primary metric, supplemented by mean absolute error (MAE). MSE is among the metrics commonly used in production forecasting~\citep{fan2025review}, and we make it primary because it gives larger errors more weight~\citep{chai2014rmse}, which matches the asymmetric economic cost of a severely inaccurate forecast; its square root, the root mean squared error, is on the same scale as the data~\citep{hyndman2006another}. MAE gives the same weight to all errors~\citep{chai2014rmse} and is less sensitive to outliers~\citep{hyndman2006another}. Reporting squared-error and absolute-error metrics together separates methods that reduce typical errors from those that specifically reduce catastrophic ones. Because \psims{} produces a predictive distribution rather than a point forecast, we additionally report the continuous ranked probability score (CRPS) computed from its sample ensemble, and, so that errors can be read in economic terms, MAE in each jurisdiction's reporting units. Comparisons on individual wells use the two-sided Wilcoxon signed-rank test on per-well errors, paired across the shared test set; $p$-values below $0.001$ are reported as $p < 0.001$.  

\paragraph{Code and data availability.} Code, configurations, and the scripts that regenerate the figures and tables of Sections~\ref{sec:clusters}--\ref{sec:ablation} and the appendices are available at \url{https://github.com/temex12/sims_ts}; the augmentation study of Section~\ref{sec:aug} is carried over from the conference version~\citep{abraha2026physics}. The evaluation artifacts (held-out targets, point forecasts, sample-trajectory stacks, and result tables) are archived at \url{https://doi.org/10.5281/zenodo.22050260}, together with a script, requiring only \texttt{numpy} and \texttt{scipy}, that recomputes 91 of the quantities reported here from the arrays and checks each against the printed value. British Columbia production data are held under a data-sharing agreement with the British Columbia Energy Regulator and are not redistributed, so the deposit covers the Alberta and Pennsylvania results; trained model weights are available from the authors on request.

\section{Results}
\label{sec:results}

\subsection{Spatial Versus Generative Augmentation: A Motivating Comparison}
\label{sec:aug}

\begin{figure}[t]
  \centering
  \includegraphics[width=0.85\textwidth]{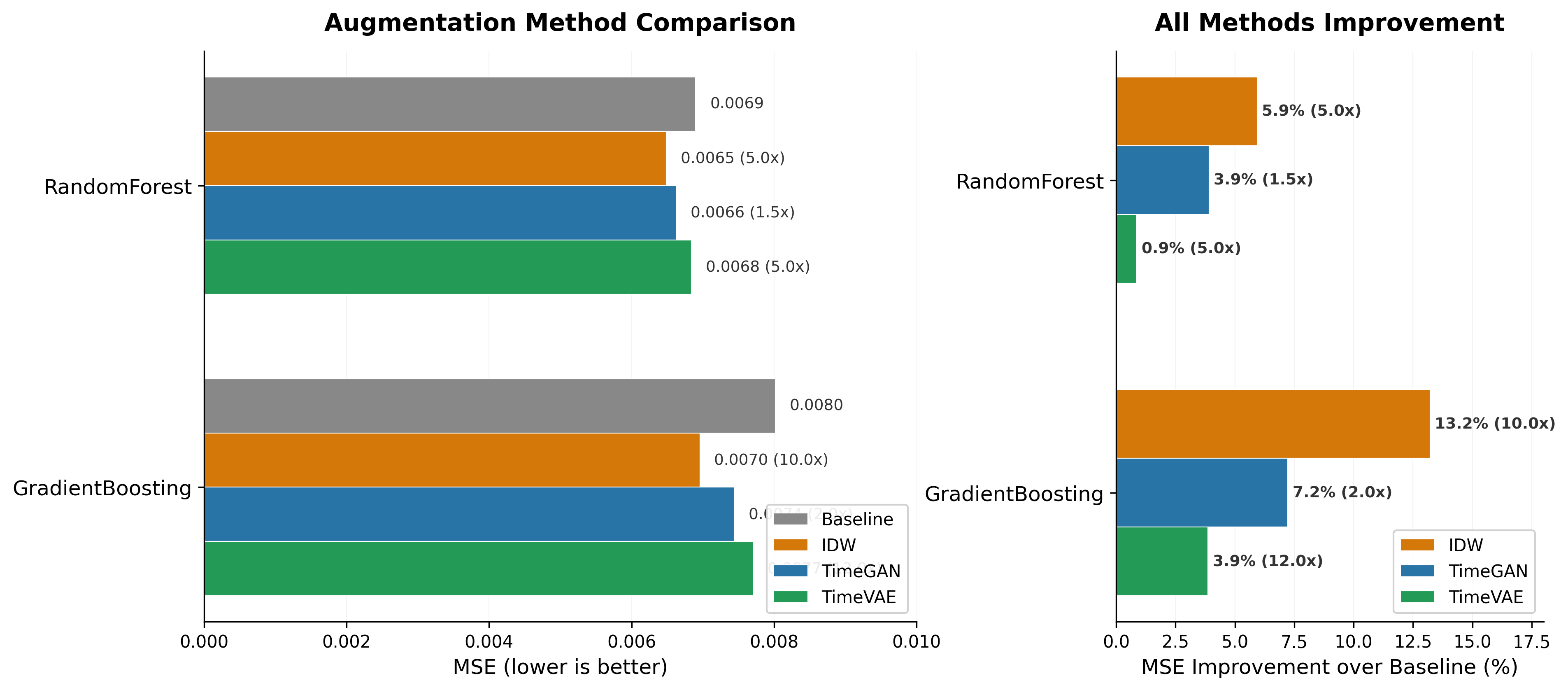}
  \caption{Data augmentation comparison on 201 Montney wells. \emph{Left:} MSE across methods for random forest and gradient boosting, with the augmentation factor in parentheses. \emph{Right:} MSE improvement over the no-augmentation baseline. IDW outperforms TimeGAN and TimeVAE for both regressors.}
  \label{fig:augmentation-mse}
\end{figure}

Before developing the full method we asked whether domain-aware augmentation offers any advantage over generic generative augmentation, using a 201-well Montney development subset (160 train, 41 test) that predates the cluster definitions of Section~\ref{sec:clusters}; this study is carried over from the conference version~\citep{abraha2026physics}. Figure~\ref{fig:augmentation-mse} and Table~\ref{tab:mse-vs-factor} report MSE across augmentation factors from 1.5 to 20 times for IDW, TimeGAN, and TimeVAE, with random forest and gradient boosting as downstream regressors. At each method's best factor, IDW achieves the largest MSE reduction for both regressors, 6.0\% for RF at the 5 times factor and 13.2\% for gradient boosting at the 10 times factor, against 4.0\% and 7.1\% for TimeGAN and 1.0\% and 3.9\% for TimeVAE; averaged over the two regressors the reductions are 9.6\%, 5.5\%, and 2.5\%. Two caveats bound what this study supports: the augmentation factor is chosen per method by minimum error on the 41-well test set, so the reductions are optimistic, and 41 test wells cannot separate improvements of this size. The comparison motivates the design that follows; the method-level claims rest on Sections~\ref{sec:clusters}--\ref{sec:ablation}.

\begin{table}[t]
\centering
\caption{MSE across augmentation factors for the random forest (RF) and gradient boosting (GB) regressors. The best result per method is shown in \emph{italics}. Baseline MSE without augmentation is 0.0069 for random forest and 0.0080 for gradient boosting. Percentage reductions reported in the text are computed from unrounded MSE.}
\label{tab:mse-vs-factor}
\small
\setlength{\tabcolsep}{4pt}
\begin{tabular}{llrrrrrrrr}
\toprule
Model & Method & 1.5$\times$ & 2$\times$ & 3$\times$ & 5$\times$ & 8$\times$ & 10$\times$ & 12$\times$ & 20$\times$ \\
\midrule
RF & IDW & 0.0071 & 0.0067 & 0.0070 & \emph{0.0065} & 0.0067 & 0.0068 & 0.0067 & 0.0066 \\
   & TimeGAN & \emph{0.0066} & 0.0068 & 0.0070 & 0.0069 & 0.0075 & 0.0071 & 0.0077 & 0.0079 \\
   & TimeVAE & 0.0069 & 0.0071 & 0.0069 & \emph{0.0068} & 0.0075 & 0.0072 & 0.0077 & 0.0077 \\
\midrule
GB & IDW & 0.0080 & 0.0078 & 0.0073 & 0.0075 & 0.0072 & \emph{0.0070} & 0.0074 & 0.0075 \\
   & TimeGAN & 0.0075 & \emph{0.0074} & 0.0077 & 0.0075 & 0.0076 & 0.0077 & 0.0078 & 0.0077 \\
   & TimeVAE & 0.0078 & 0.0078 & 0.0079 & 0.0077 & 0.0078 & 0.0078 & \emph{0.0077} & 0.0080 \\
\bottomrule
\end{tabular}
\end{table}

Table~\ref{tab:mse-vs-factor} shows that IDW stays stable across factors, with its optimum between 5 and 10 times, whereas TimeGAN and TimeVAE degrade as the factor grows; TimeGAN MSE rises from 0.0066 at 1.5 times to 0.0079 at 20 times for RF. The generative augmenters appear to introduce increasingly unrealistic samples at high factors, while IDW interpolation preserves the structure of real sequences. Gradient boosting benefits more from augmentation than random forest across all methods, consistent with its sequential error-correction exploiting the added samples. The stability of IDW across factors, rather than the size of any single margin, is what motivated the design that follows: encoding domain knowledge directly, both as the spatial augmentation retained in the training recipe of Section~\ref{sec:psims-config} and as the decline-curve constraints of Section~\ref{sec:physics}.

\subsection{Production Forecasting Across Formation Clusters}
\label{sec:clusters}

\begin{table}[t]
\centering
\caption{Mean squared error across the British Columbia formation clusters (per-curve space, frozen configuration of Section~\ref{sec:protocol}); every column is a five-instance ensemble. Clusters below 2{,}000 wells use members fine-tuned from pretraining models whose training wells exclude every cluster's test wells. The two PatchTST columns are the same architecture trained on the formation alone and given the identical pretrain-and-fine-tune pipeline. The best value per row is shown in \emph{italics}.}
\label{tab:cluster-results}
\small
\setlength{\tabcolsep}{4pt}
\begin{tabular}{l r r r r r rr}
\toprule
 & & & & & & \multicolumn{2}{c}{PatchTST} \\
\cmidrule(lr){7-8}
Formation & Wells & Test & IDW+RF & \simsts{} & \psims{} & scratch & pretrained \\
\midrule
Wilrich      &     51 &   11 & 0.0098 & 0.0067 & \emph{0.0066} & 0.0158 & 0.0075 \\
Golata       &     53 &   11 & 0.0380 & \emph{0.0271} & 0.0299 & 0.0404 & 0.0278 \\
Debolt       &    188 &   36 & \emph{0.0222} & 0.0230 & 0.0236 & 0.0246 & 0.0224 \\
Baldonnel    &    293 &   58 & 0.0227 & 0.0209 & 0.0218 & 0.0237 & \emph{0.0199} \\
Charlie Lake &    690 &  138 & 0.0205 & 0.0180 & 0.0181 & 0.0192 & \emph{0.0168} \\
Doig         &  1{,}390 &  276 & 0.0151 & 0.0145 & 0.0146 & 0.0151 & \emph{0.0138} \\
Montney      &  5{,}819 & 1{,}163 & 0.0123 & 0.0123 & 0.0124 & 0.0123 & \emph{0.0120} \\
All Wells    & 16{,}216 & 3{,}244 & 0.0172 & 0.0171 & 0.0171 & \multicolumn{2}{c}{\emph{0.0162}} \\
\bottomrule
\end{tabular}
\end{table}

Table~\ref{tab:cluster-results} reports the formation clusters, which span 51 to 16{,}216 wells and are the setting in which one would expect a generative prior to help most. The diffusion forecasters attain a lower MSE than IDW+RF on six of the eight clusters, with the largest margins on the smallest formations (Wilrich $-32\%$, Golata $-21\%$), and they improve substantially on a PatchTST trained on the formation alone, by 58\% on Wilrich and 26\% on Golata. The two exceptions are Debolt and Montney, where the spatial baseline is ahead of the better diffusion variant by 3.7\% and 0.2\%.

That last comparison, however, measures the pipeline rather than the model, and we report the control that separates them. \psims{} reaches the formations below 2{,}000 wells by pretraining on all other British Columbia wells and fine-tuning. A PatchTST given the same pipeline attains the lower error on six of the seven formations, and on five of the six where \psims{} is also pretrained (Montney, at 5{,}819 wells, is above the fine-tuning threshold, so \psims{} trains there from scratch). Wilrich is the exception: \psims{} remains ahead of the pretrained PatchTST on all eleven test wells ($p = 0.001$), but a pretrained iTransformer reaches 0.0070 there and the difference against it is not significant ($p = 0.15$), and eleven test wells are too few to support a small-data claim either way. The advantage on small formations therefore belongs to cross-formation pretraining, which is available to any architecture, and not to the diffusion model.

\begin{figure}[t]
  \centering
  \includegraphics[width=\textwidth]{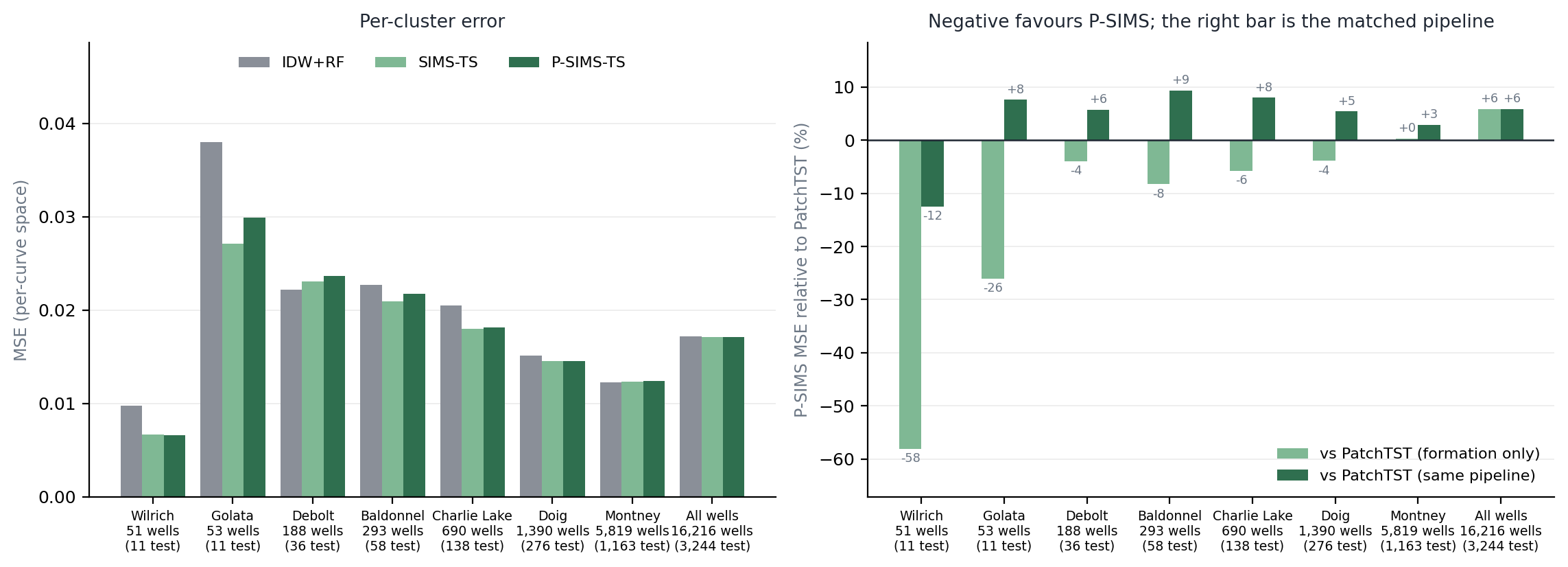}
  \caption{Formation clusters, ordered by size. \emph{Left:} MSE per method per cluster. \emph{Right:} \psims{} relative to PatchTST at matched training budget, with negative values favouring \psims{}. The light bars compare against a PatchTST trained on the formation alone and show a large apparent advantage on the smallest formations; the dark bars give PatchTST the same pretrain-and-fine-tune pipeline that \psims{} uses, and the advantage disappears on six of the seven formations. The all-wells row is the full dataset, where the two treatments coincide.}
  \label{fig:cluster-comparison}
\end{figure}

Figure~\ref{fig:cluster-comparison} shows the same picture per cluster. \simsts{} and \psims{} agree within 1\% on five of the eight clusters, including every cluster with more than 100 test wells, and differ on Golata, Baldonnel, and Debolt by 10.2\%, 4.2\%, and 2.6\% of MSE with \simsts{} ahead; those three clusters have 11, 58, and 36 test wells. We do not read a method-level effect into differences of that size on so few wells, since the split noise at these test sizes exceeds the effect and the equally small Wilrich set places \psims{} 0.8\% ahead instead. Section~\ref{sec:ablation}, which measures the physics contribution on thousands of wells at a time under matched sampling, is the evidence we rely on. What \psims{} adds on every cluster, small or large, is that its forecasts are monotone by construction.

\subsection{Cross-Jurisdiction Forecasting Accuracy}
\label{sec:leaderboard}

\begin{table}[t]
\centering
\caption{Cross-jurisdiction results on the all-wells cluster of each jurisdiction, all methods trained and evaluated in the shared per-curve space. The upper block is the matched comparison every claim in this article is stated against: five-instance ensembles trained for up to 200 epochs at patience 30. The lower block reports the remaining baselines as single trained models under their fixed configurations; Table~\ref{tab:diffusion-family} compares the diffusion baselines against a single \psims{} model. The best value per column within each block is shown in \emph{italics}. BC denotes British Columbia and PA Marc.\ Pennsylvania Marcellus.}
\label{tab:jurisdictions}
\footnotesize
\setlength{\tabcolsep}{3.2pt}
\begin{tabular}{l rr rr rr}
\toprule
 & \multicolumn{2}{c}{BC (16{,}216)} & \multicolumn{2}{c}{Alberta (6{,}747)} & \multicolumn{2}{c}{PA Marc.\ (12{,}360)} \\
\cmidrule(lr){2-3}\cmidrule(lr){4-5}\cmidrule(lr){6-7}
Method & MSE & MAE & MSE & MAE & MSE & MAE \\
\midrule
\multicolumn{7}{@{}l}{\makebox[0pt][l]{\textit{Five-instance ensembles at matched training budget:}}} \\
\psims{} (ours) & 0.0171 & 0.0851 & 0.0172 & \emph{0.0855} & 0.00916 & \emph{0.0623} \\
IDW+RF          & 0.0172 & 0.0879 & 0.0173 & 0.0885 & 0.00917 & 0.0633 \\
PatchTST        & \emph{0.0162} & \emph{0.0847} & \emph{0.0163} & 0.0856 & \emph{0.00910} & 0.0630 \\
iTransformer    & 0.0169 & 0.0860 & 0.0172 & 0.0878 & 0.00966 & 0.0650 \\
\midrule
\multicolumn{7}{@{}l}{\makebox[0pt][l]{\textit{Single-configuration baselines, one trained model each:}}} \\
CSDI            & \emph{0.0191} & \emph{0.0912} & \emph{0.0232} & \emph{0.0989} & \emph{0.01185} & \emph{0.0709} \\
Diffusion-TS    & 0.0280 & 0.1073 & 0.0254 & 0.1058 & 0.01521 & 0.0816 \\
TSDiff          & 0.0280 & 0.1068 & 0.0301 & 0.1228 & 0.01881 & 0.1043 \\
TimeGrad        & 0.0350 & 0.1200 & 0.0330 & 0.1191 & 0.01886 & 0.0907 \\
Chronos-2       & 0.0704 & 0.2046 & 0.0738 & 0.2149 & 0.09229 & 0.2528 \\
Chronos-Bolt    & 0.1138 & 0.2792 & 0.1132 & 0.2800 & 0.13346 & 0.3099 \\
Arps (prefix fit) & 0.0503 & 0.1494 & 0.0563 & 0.1645 & 0.05190 & 0.1553 \\
\bottomrule
\end{tabular}
\end{table}

\begin{figure}[t]
  \centering
  \includegraphics[width=\textwidth]{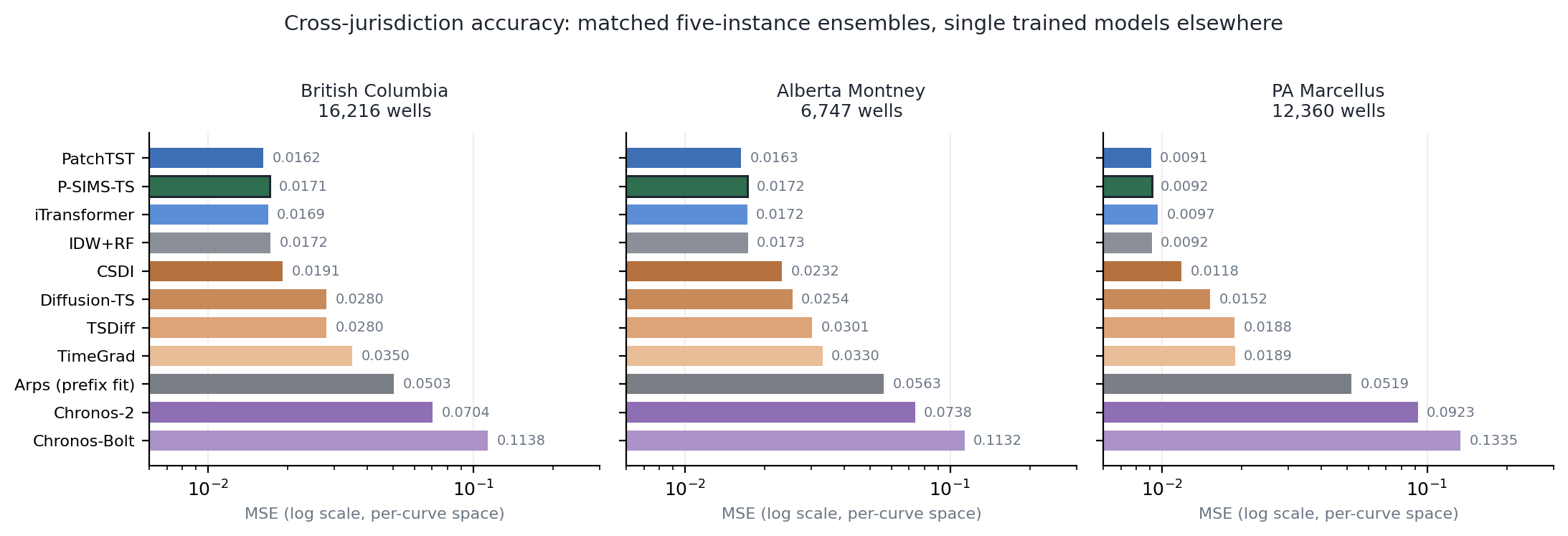}
  \caption{Cross-jurisdiction MSE on the all-wells cluster, logarithmic scale, showing Table~\ref{tab:jurisdictions}: \psims{}, IDW+RF, PatchTST, and iTransformer at the matched five-instance budget, and the remaining baselines as single trained models. Methods are ordered by mean rank across the three jurisdictions, so the order is shared by the panels. Ensembled PatchTST attains the lowest mean squared error on British Columbia and Alberta, and its Pennsylvania margin over \psims{} is not resolved; the ordering of the baselines below them is nearly identical across jurisdictions. The classical Arps prefix fit trails every method that learns across wells while remaining ahead of both zero-shot foundation models.}
  \label{fig:cross-jurisdiction}
\end{figure}

Table~\ref{tab:jurisdictions} reports the leaderboard and Figure~\ref{fig:cross-jurisdiction} visualizes it. The last row is the domain's classical method: the Arps hyperbolic model of \eqref{eq:arps} fitted by bounded least squares to each well's own 20-point prefix and extrapolated over the horizon, with no information shared across wells. It converges on every test well and is monotone by construction, though the fit is constrained by a parameter bound on at least three quarters of wells and returns a near-flat extrapolation on a fifth to a quarter of them, which is the practical form its failure on a short prefix takes. It trails every method that learns across wells, at 2.9 to 5.7 times the MSE of \psims{}, which is more accurate on 76.7\% of British Columbia wells, 76.6\% of Alberta wells, and 82.7\% of Pennsylvania wells ($p < 0.001$ in each jurisdiction); it is nevertheless more accurate than either zero-shot foundation model. Fitting the same model on the physical rate rather than in the normalized space and mapping the result back gives 0.0508, 0.0567, and 0.0525, within about 1\% of the per-curve fit, so the fitting space is not what places this baseline last. This quantifies the claim made in the introduction: decline-curve fitting on a short prefix is not competitive at this forecast origin, which is what motivates learning across wells.

Two readings of Table~\ref{tab:jurisdictions} matter, and they are different claims. Within the diffusion family, \psims{} is the strongest method, and Section~\ref{sec:diffusion-family} establishes that at the level of individual wells and at matched budgets. Against the field as a whole it is \emph{competitive}, not superior. Under a matched five-instance budget, ensembled PatchTST attains a lower mean squared error in all three jurisdictions, by 5.5\% on British Columbia and 5.4\% on Alberta, both of which resolve under a paired bootstrap over wells. The Pennsylvania margin is 0.7\% and does not: its 95\% interval spans 1.5\% in \psims{}'s favour to 2.9\% in PatchTST's, and the paired per-well test on the same arrays resolves for \psims{} (Table~\ref{tab:significance}), so the Pennsylvania comparison is unresolved rather than won by either method. On mean absolute error the two are level: \psims{} leads on Alberta (0.0855 against 0.0856) and Pennsylvania (0.0623 against 0.0630) and trails on British Columbia (0.0851 against 0.0847), with none of the three differences large. Against the other two matched-budget baselines the picture is mixed on squared error and consistent on absolute error: \psims{} improves on ensembled IDW+RF in every jurisdiction on both metrics, and on ensembled iTransformer on mean absolute error in every jurisdiction, while ensembled iTransformer attains the lower mean squared error on British Columbia (0.0169 against 0.0171) and Alberta (0.017195 against 0.017225). Every number in this section uses the matched 200-epoch budget; Section~\ref{sec:protocol-effects} reports how the transformer comparison changes under the shorter size-adaptive schedule of the baseline implementations.

\begin{table}[t]
\centering
\caption{Paired per-well comparison of \psims{} against the matched-budget baseline ensembles, all trained for 200 epochs. Win rate is the fraction of individual wells on which \psims{} attains the lower error; $p$ is a two-sided Wilcoxon signed-rank test on per-well differences and is therefore undirected. Entries that do not resolve at the 5\% level are marked $\circ$; entries that resolve \emph{against} \psims{} are marked $\blacktriangledown$.}
\label{tab:significance}
\footnotesize
\begin{tabular}{ll rr rr}
\toprule
 & & \multicolumn{2}{c}{squared error} & \multicolumn{2}{c}{absolute error} \\
\cmidrule(lr){3-4}\cmidrule(lr){5-6}
 & Baseline (5-instance) & win & $p$ & win & $p$ \\
\midrule
BC & IDW+RF & 54.2\% & $<0.001$ & 55.9\% & $<0.001$ \\
   & PatchTST & 50.1\% & $0.015^{\,\blacktriangledown}$ & 51.7\% & $0.34^{\,\circ}$ \\
   & iTransformer & 52.6\% & $0.43^{\,\circ}$ & 54.7\% & $<0.001$ \\
\midrule
Alberta & IDW+RF & 53.6\% & $0.010$ & 53.9\% & $<0.001$ \\
        & PatchTST & 51.9\% & $0.65^{\,\circ}$ & 52.5\% & $0.057^{\,\circ}$ \\
        & iTransformer & 55.3\% & $0.001$ & 55.8\% & $<0.001$ \\
\midrule
PA & IDW+RF & 51.3\% & $0.47^{\,\circ}$ & 51.6\% & $0.015$ \\
   & PatchTST & 53.4\% & $0.021$ & 53.9\% & $<0.001$ \\
   & iTransformer & 56.4\% & $<0.001$ & 57.2\% & $<0.001$ \\
\bottomrule
\end{tabular}
\end{table}

Against the single-configuration diffusion and foundation baselines the margins are large and decisive in all three jurisdictions: every paired comparison gives Wilcoxon $p < 0.001$ with \psims{} win rates of 64--98\%. In British Columbia specifically, \psims{} is more accurate than CSDI on 64.2\% of wells, Diffusion-TS on 73.8\%, TSDiff on 72.6\%, TimeGrad on 76.0\%, Chronos-2 on 90.0\%, and Chronos-Bolt on 93.3\%. Section~\ref{sec:diffusion-family} repeats the diffusion comparison at matched training and sampling budgets, where it also holds.

Against the matched-budget ensembles the picture is closer. \psims{} attains the lower error on a majority of wells in all eighteen comparisons of Table~\ref{tab:significance}, but the majorities are small against ensembled PatchTST (50.1--53.9\%), five of the eighteen tests do not resolve at the 5\% level, and one, British Columbia against PatchTST on squared error, resolves in the baseline's favour. No multiplicity correction is applied; at a Bonferroni-corrected threshold for eighteen tests ($\alpha = 0.0028$) nine remain significant. Against ensembled PatchTST the two methods are close enough that per-well ordering does not separate them consistently: Pennsylvania resolves for \psims{} on both metrics, British Columbia resolves against it on squared error, and the remainder do not resolve. \psims{} is ahead of ensembled IDW+RF on both metrics everywhere and ahead of ensembled iTransformer on absolute error everywhere.

\paragraph{Probabilistic forecasts and data-unit errors.}
\psims{} retains its full sample ensemble, so a distributional metric is available for it: its CRPS over the 100 trajectories is 0.047 (Pennsylvania), 0.066 (Alberta), and 0.065 (British Columbia) on the normalized scale. For a point forecast the CRPS reduces to the MAE, so the baselines, which are evaluated as point forecasts, stand at the MAE values of Table~\ref{tab:jurisdictions} on this scale. The \psims{} point forecast attains a mean absolute error of 0.0623 on Pennsylvania against its CRPS of 0.0473, so the difference between the two measures what the predictive spread adds over the forecast centre rather than a more accurate centre.

The predictive distribution is overconfident before calibration and usable after it. Taking central intervals across the 100 trajectories, nominal 50\% intervals cover 30--32\% of held-out observations, nominal 80\% cover 53--56\%, and nominal 90\% cover 64--67\%. Under-dispersion is a known property of deep ensembles and does not affect the CRPS values above. It is also easily corrected: inflating the ensemble spread about its mean by a single scalar, fitted on half the test wells and evaluated on the other half, restores coverage to 53--54\% at the 50\% level, 81--83\% at the 80\% level, and 89--90\% at the 90\% level. One factor per jurisdiction is enough, 1.80 for British Columbia, 1.90 for Alberta, and 1.80 for Pennsylvania, each fitted at the 90\% level on the fitting half and applied unchanged to the other two levels; on a different half-split the factors move by about 0.05, which does not change the coverage they restore. One parameter therefore turns the trajectory ensemble into calibrated intervals, and because the samples are joint curves rather than pointwise bands, the same ensemble yields a distribution over any functional of the whole forecast. A five-trajectory subset with its own factor also reaches 89--90\% at the 90\% level but over-covers at the other two levels (60--61\% and 86--88\% against nominal 50\% and 80\%), whereas one factor calibrates the full ensemble at all three levels at once. In the reporting units of each jurisdiction, the mean absolute forecast error of \psims{} is 12{,}623 Mcf/month on Pennsylvania (median well: 7{,}347), 158 $10^3$m$^3$/month on Alberta, and 201 $10^3$m$^3$/month on British Columbia, against 12{,}925 and 12{,}744 Mcf/month for ensembled IDW+RF and ensembled PatchTST on Pennsylvania, 165 and 162 on Alberta, and 207 and 200 on British Columbia. In data units the matched-budget methods therefore lie within about five percent of one another in every jurisdiction, which is the practical size of the differences this comparison resolves.

\paragraph{Computational cost.}
Table~\ref{tab:padep} reports compute cost on Pennsylvania, the largest evaluation. \psims{} is the most expensive method in the comparison: 203 minutes to build and train the five ensemble members and 40 minutes to draw and post-process their 100 trajectories over 2{,}472 test wells, against 45--52 minutes for the heaviest diffusion baselines and 2 to 4 minutes per instance for the transformers. Most of the premium is the ensemble: a single member with $K = 20$ samples costs 41 minutes to train and 8 to sample, of which 4.4 minutes is the spatial augmentation of its training set. Within sampling, the physics terms are not free: on a given member, an unguided, unprojected pass over the test set takes 4.7 minutes against 8.8 for the guided and projected one, so the constraints roughly double sampling cost, although sampling is itself under a fifth of the total. At the same five-instance budget and the same 200 epochs, ensembled PatchTST costs 20.3 minutes and ensembled iTransformer 11.4, so \psims{} is roughly twelve to twenty-one times more expensive than the transformer baselines it does not outperform. IDW+RF is the second most expensive method at 43.4 minutes, because its spatial augmentation dominates the downstream regressor fit; all times are end-to-end for the full method rather than for the final estimator alone. Inference cost scales linearly in $R \times K$, which is the parameter a deployment would tune first; the compute buys the predictive distribution described above rather than accuracy.

\begin{table}[t]
\centering
\caption{Compute cost on Pennsylvania Marcellus (12{,}360 wells; 9{,}888 train / 2{,}472 test), single-GPU wall clock on an RTX 6000 Ada. The middle column is training plus inference for one trained instance at that method's sampling budget of Section~\ref{sec:protocol}; the last column gives the cost of the five-instance configurations compared in Table~\ref{tab:jurisdictions}, where those were run. The Chronos models are applied zero-shot, so their cost is inference only.}
\label{tab:padep}
\small
\begin{tabular}{lrr}
\toprule
Method & One instance & Five instances \\
\midrule
\psims{} (ours)     & 48.7\,min & 243\,min \\
IDW+RF              & 8.7\,min  & 43.4\,min \\
PatchTST            & 4.1\,min  & 20.3\,min \\
iTransformer        & 2.3\,min  & 11.4\,min \\
CSDI                & 6.1\,min  & -- \\
Diffusion-TS        & 15.7\,min & -- \\
TSDiff              & 45.1\,min & -- \\
TimeGrad            & 51.8\,min & -- \\
Chronos-2           & 21.6\,min & -- \\
Chronos-Bolt        & 2.7\,s    & -- \\
\bottomrule
\end{tabular}
\end{table}

\subsection{Comparison Within the Diffusion Family}
\label{sec:diffusion-family}

Each of CSDI, Diffusion-TS, TSDiff, and TimeGrad trains under its own published size-adaptive schedule, which on these datasets allows 200 epochs for CSDI and TimeGrad and 300 for Diffusion-TS and TSDiff, against 200 for \psims{}. No method reaches its cap: on British Columbia early stopping fires at 50 epochs for TSDiff, 51 for TimeGrad, 52 for CSDI, and 102 for Diffusion-TS, against 126 for the \psims{} member reported here, so the budgets bind neither side. The training-set asymmetry of Section~\ref{sec:protocol} applies here as well. Each baseline draws 10 stochastic samples per well and reports the per-timestep median. To remove the ensembling advantage as well, Table~\ref{tab:diffusion-family} compares them against a \emph{single} \psims{} model rather than the $R{=}5$ ensemble used elsewhere, so both sides are one trained model decoded from multiple samples.

\begin{table}[t]
\centering
\caption{Diffusion forecasters at matched budgets: one trained model each, multi-sample decoding, evaluated on the shared test wells. The \psims{} column is a single ensemble member, not the $R{=}5$ ensemble reported elsewhere. Win rate is the fraction of individual wells on which \psims{} attains the lower squared error; $p$ is a two-sided Wilcoxon signed-rank test on per-well differences. The best value per column is shown in \emph{italics}.}
\label{tab:diffusion-family}
\footnotesize
\setlength{\tabcolsep}{2.8pt}
\begin{tabular}{l rr r rr r rr r}
\toprule
 & \multicolumn{3}{c}{BC (3{,}244 test)} & \multicolumn{3}{c}{Alberta (1{,}350 test)} & \multicolumn{3}{c}{PA Marcellus (2{,}472 test)} \\
\cmidrule(lr){2-4}\cmidrule(lr){5-7}\cmidrule(lr){8-10}
Method & MSE & MAE & win & MSE & MAE & win & MSE & MAE & win \\
\midrule
\psims{} (one model) & \emph{0.0181} & \emph{0.0875} & -- & \emph{0.0177} & \emph{0.0873} & -- & \emph{0.0096} & \emph{0.0636} & -- \\
CSDI                 & 0.0191 & 0.0912 & 59.9\% & 0.0232 & 0.0989 & 69.1\% & 0.0118 & 0.0709 & 66.6\% \\
Diffusion-TS         & 0.0280 & 0.1073 & 71.3\% & 0.0254 & 0.1058 & 72.1\% & 0.0152 & 0.0816 & 71.4\% \\
TSDiff               & 0.0280 & 0.1068 & 69.9\% & 0.0301 & 0.1228 & 78.8\% & 0.0188 & 0.1043 & 82.2\% \\
TimeGrad             & 0.0350 & 0.1200 & 75.0\% & 0.0330 & 0.1191 & 76.3\% & 0.0189 & 0.0907 & 76.1\% \\
\bottomrule
\end{tabular}
\end{table}

\psims{} is more accurate than every diffusion baseline on both metrics in all three jurisdictions, twelve of twelve comparisons, and the margin is not carried by a subset of wells: it attains the lower squared error on 60 to 82\% of individual wells, with every paired test at $p < 0.001$. The closest competitor is CSDI, whose conditional formulation is nearest to our backbone; the gap to it is 5\% on British Columbia and 19 to 24\% on the other two jurisdictions.

One asymmetry remains: the baselines draw 10 samples and take the per-timestep median, while \psims{} draws 20 samples, each projected at the end of sampling, averages them, and applies the cumulative minimum. Decoding a single \psims{} model under the baselines' rule (10 samples, median) gives 0.0188 (British Columbia), 0.0184 (Alberta), and 0.0098 (Pennsylvania), narrowing the margin over CSDI on British Columbia from 5\% to 2.0\%. Removing the projection as well, so that neither side applies any shape post-processing, gives 0.0191, 0.0184, and 0.0098: Alberta and Pennsylvania are unaffected at 20.9\% and 16.9\%, but the British Columbia margin falls to 0.5\%. The ordering holds in all twelve comparisons under every decoding rule, and the projection is available to the baselines at the cost measured in Table~\ref{tab:ablation-baselines}, but the margin over CSDI on British Columbia is small enough that the family-level claim rests on Alberta and Pennsylvania. The $R{=}5$ ensemble of Table~\ref{tab:jurisdictions} widens each of these margins further, and the same ordering holds on the six standard benchmarks of Section~\ref{sec:benchmarks}, where the reversible-instance-normalization variant of the backbone leads the diffusion baselines across three seeds.

We attribute this standing to the training and inference recipe of Section~\ref{sec:psims-config} rather than to the physics, which Section~\ref{sec:ablation} shows to be accuracy-neutral: the shared per-curve representation, the spatial training augmentation, the exponential-moving-average weights with min-SNR weighting, and the trajectory ensemble are what make a diffusion forecaster competitive with the transformer frontier here rather than trailing it by the factors of 1.2 to 2.2 that the untuned diffusion baselines show.

\subsection{Cross-Jurisdiction Transfer}
\label{sec:transfer}

The leaderboard of Section~\ref{sec:leaderboard} trains and tests within a single jurisdiction. We next apply each trained model, without retraining, to the held-out test wells of the other two. All three models are trained under the matched 200-epoch budget of Section~\ref{sec:protocol}; \psims{} uses its five-instance ensemble carrying its own formation-fitted prior, and the transformers are single instances.

\begin{table}[t]
\centering
\small
\setlength{\tabcolsep}{4pt}
\begin{tabular}{llrrr}
\toprule
Model & Train $\downarrow$ / Test $\rightarrow$ & BC & Alberta & PA Marc. \\
\midrule
\psims{}      & BC           & \emph{0.0171} & 0.0173 ($+0.0000$) & 0.0099 ($+0.0008$) \\
              & Alberta      & 0.0179 ($+0.0007$) & \emph{0.0172} & 0.0100 ($+0.0008$) \\
              & PA Marcellus & 0.0180 ($+0.0009$) & 0.0178 ($+0.0006$) & \emph{0.0092} \\
\midrule
PatchTST      & BC           & \emph{0.0163} & 0.0163 ($-0.0001$) & 0.0098 ($+0.0005$) \\
              & Alberta      & 0.0170 ($+0.0007$) & \emph{0.0164} & 0.0100 ($+0.0006$) \\
              & PA Marcellus & 0.0174 ($+0.0011$) & 0.0172 ($+0.0009$) & \emph{0.0093} \\
\midrule
iTransformer  & BC           & \emph{0.0171} & 0.0174 ($+0.0001$) & 0.0099 ($+0.0002$) \\
              & Alberta      & 0.0178 ($+0.0007$) & \emph{0.0173} & 0.0107 ($+0.0010$) \\
              & PA Marcellus & 0.0179 ($+0.0008$) & 0.0177 ($+0.0004$) & \emph{0.0097} \\
\bottomrule
\end{tabular}
\caption{Cross-jurisdiction transfer MSE at matched training budget. Diagonal entries, shown in \emph{italics}, are within-jurisdiction results for that model. Parenthesized deltas compare each transfer cell against the \emph{target} jurisdiction's own model of the same family, so both quantities share a test set. All \psims{} forecasts are monotone by construction; the transformer forecasts are not.}
\label{tab:transfer}
\end{table}

Table~\ref{tab:transfer} reports the three matrices, and two observations follow. First, transfer degradation is small for every model and smallest for \psims{}: applying a jurisdiction's model to another costs \psims{} at most $+0.0009$ MSE relative to the target's own model, against $+0.0010$ for iTransformer and $+0.0011$ for PatchTST. The ordering is consistent, but the differences are within a tenth of the between-model gaps, so the fair statement is that all three families transfer comparably well between these plays, including between the Montney jurisdictions and the geologically distinct Marcellus. Second, absolute accuracy under transfer follows the within-jurisdiction ranking rather than reversing it. A PatchTST trained on British Columbia forecasts Pennsylvania at 0.0098, better than either transferred \psims{} model (0.0099 and 0.0100) though not better than \psims{} trained on Pennsylvania itself (0.0092), and Pennsylvania's own transformer baselines at the matched budget (0.0093 and 0.0097) are ahead of the transferred \psims{} models as well. What the transfer study establishes is that the decline prior does not bind \psims{} to its source geology, not that it confers an accuracy advantage away from home; Section~\ref{sec:protocol-effects} reports how this comparison depends on the training budget.

\subsection{Leave-One-Domain-Out Generalization}
\label{sec:lodo}

A more demanding test pools the remaining jurisdictions: each model trains on the combined training wells of two jurisdictions and predicts the third, which it has never seen.

\begin{table}[t]
\centering
\small
\begin{tabular}{lrrr}
\toprule
Model & $\to$BC & $\to$Alberta & $\to$PA Marc. \\
 & (Alberta+PA) & (BC+PA) & (BC+Alberta) \\
\midrule
\psims{}     & 0.0177 & 0.0172 & 0.0099 \\
iTransformer & 0.0175 & 0.0170 & 0.0097 \\
PatchTST     & \emph{0.0167} & \emph{0.0163} & \emph{0.0096} \\
\bottomrule
\end{tabular}
\caption{Leave-one-domain-out MSE at matched training budget: each model trains on the pooled training wells of two jurisdictions and predicts the held-out jurisdiction's test wells. The best value per fold is shown in \emph{italics}. \psims{} forecasts are monotone on every fold.}
\label{tab:lodo}
\end{table}

Table~\ref{tab:lodo} reports the held-out MSE. PatchTST attains the lowest error on all three folds and \psims{} the highest, with iTransformer between them; the spread is 3 to 6\%. The result is consistent with the within-jurisdiction result of Table~\ref{tab:jurisdictions}: the pooled-training setting does not change the relative standing of the three families (Section~\ref{sec:protocol-effects} reports the same folds under the shorter training schedule).

The quantity that does not degrade is the gap between pooled and within-jurisdiction training. Pooled training costs \psims{} $+0.0005$ on British Columbia and $+0.0008$ on Pennsylvania relative to its own five-instance within-jurisdiction ensembles, and improves its Alberta result by $0.0001$. PatchTST's leave-one-domain-out models are single instances, and against its single-instance within-jurisdiction results its pooled training costs $+0.0004$ on British Columbia, $-0.0000$ on Alberta, and $+0.0002$ on Pennsylvania. For Alberta, the jurisdiction with the least training data, a model trained without a single Alberta well matches one trained on Alberta for both families, which is a statement about the transferability of decline behaviour across these plays rather than about either architecture.

\subsection{Standard-Benchmark Competitiveness of the Backbone}
\label{sec:benchmarks}

\begin{figure}[t]
  \centering
  \includegraphics[width=0.85\textwidth]{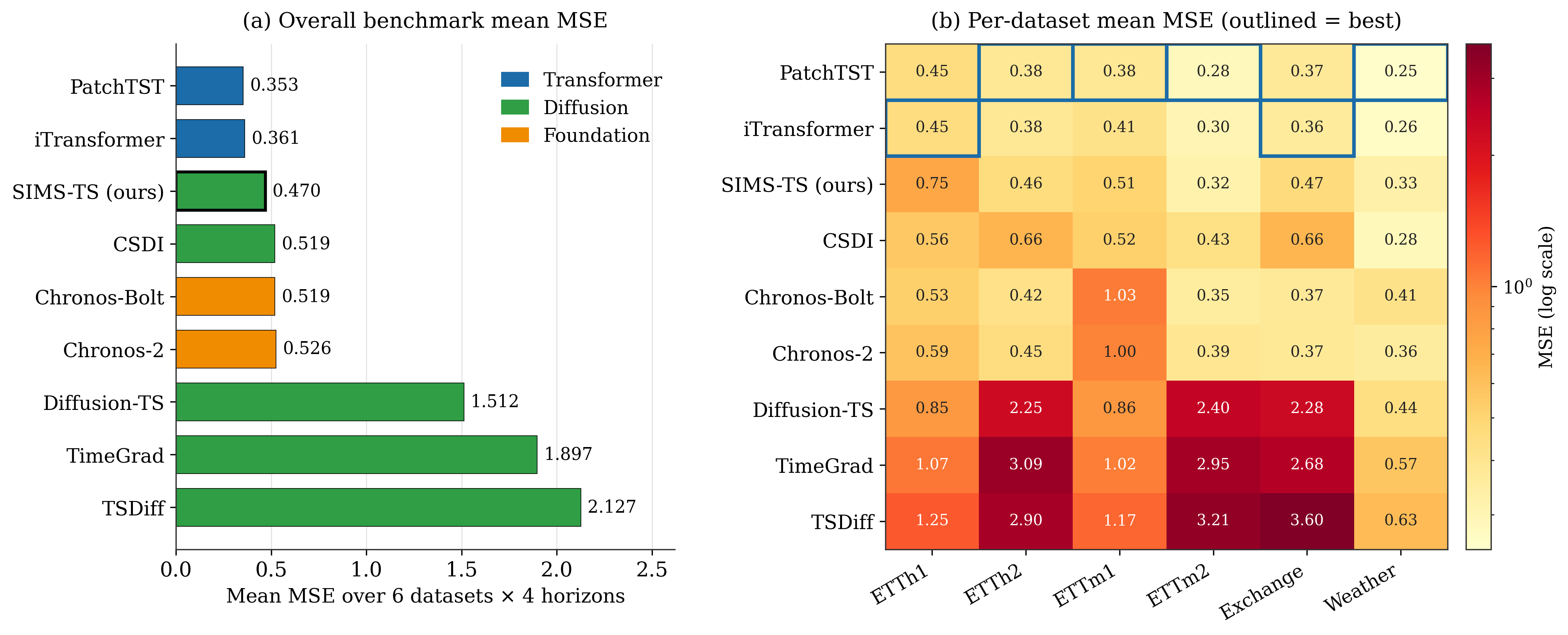}
  \caption{Standard benchmark suite. \emph{(a)} Overall mean MSE per model, colored by family; the RevIN \simsts{} variant is the leading diffusion baseline, behind the two transformers. \emph{(b)} Per-dataset mean MSE on a logarithmic scale, with the per-dataset best outlined. The diffusion baselines Diffusion-TS, TimeGrad, and TSDiff are unstable on the non-stationary datasets.}
  \label{fig:benchmark}
\end{figure}

A natural question is whether the diffusion backbone is competitive outside the production domain. We evaluate the RevIN variant of Section~\ref{sec:revin} on the six standard datasets against the same baselines, without the production-specific physics guidance and with $\omega$ fixed at 0 (Section~\ref{sec:impl}). Table~\ref{tab:bench-overall} reports the per-dataset and overall mean MSE, and Figure~\ref{fig:benchmark} summarizes both views. The RevIN \simsts{} variant reaches an overall mean MSE of 0.471, ahead of CSDI at 0.519, which makes it the leading diffusion baseline on the suite. It produces a lower MSE than CSDI on 13 of the 24 settings, concentrated on the non-stationary datasets: every horizon of ETTm2 and Exchange, horizons 192, 336, and 720 of ETTh2, the 720 horizon of ETTm1, and the 96 horizon of Weather. CSDI keeps the lead on ETTh1 and on the longer Weather horizons. The transformers PatchTST and iTransformer remain ahead overall, at 0.353 and 0.361, so the diffusion backbone leads its own family but does not reach the transformer frontier. The purpose of this study is narrower than winning the benchmark: it establishes that the backbone the production pipeline builds on is not itself the limiting factor, so the production results of Sections~\ref{sec:clusters} and~\ref{sec:leaderboard} do not rest on a base model that is weak by construction. Per-horizon results for all models appear in Appendix~\ref{app:bench}, and Section~\ref{sec:multiseed} repeats the evaluation over three random seeds for the six most closely ranked models.

\begin{table}[t]
\centering
\caption{Per-dataset and overall mean MSE across the four horizons of the standard benchmark suite. The best value per column is shown in \emph{italics}. The RevIN \simsts{} variant is the leading diffusion baseline overall, driven by the non-stationary datasets ETTh2, ETTm2, and Exchange.}
\label{tab:bench-overall}
\small
\setlength{\tabcolsep}{5pt}
\resizebox{\textwidth}{!}{\begin{tabular}{cl rrrrrrr}
\toprule
Rank & Model & ETTh1 & ETTh2 & ETTm1 & ETTm2 & Exchange & Weather & Overall \\
\midrule
1 & PatchTST       & 0.451          & \emph{0.379} & \emph{0.383} & \emph{0.283} & 0.371        & \emph{0.253} & \emph{0.353} \\
2 & iTransformer   & \emph{0.450}   & 0.380        & 0.413        & 0.302        & \emph{0.361} & 0.261        & 0.361 \\
3 & \simsts{} (ours) & 0.750        & 0.459        & 0.506        & 0.317        & 0.469        & 0.325        & 0.471 \\
4 & CSDI           & 0.558          & 0.661        & 0.519        & 0.432        & 0.658        & 0.285        & 0.519 \\
5 & Chronos-Bolt   & 0.525          & 0.425        & 1.033        & 0.352        & 0.368        & 0.413        & 0.520 \\
6 & Chronos-2      & 0.587          & 0.452        & 0.998        & 0.390        & 0.374        & 0.356        & 0.526 \\
7 & Diffusion-TS   & 0.846          & 2.251        & 0.857        & 2.403        & 2.278        & 0.440        & 1.513 \\
8 & TimeGrad       & 1.071          & 3.091        & 1.022        & 2.946        & 2.683        & 0.571        & 1.897 \\
9 & TSDiff         & 1.252          & 2.902        & 1.165        & 3.211        & 3.603        & 0.627        & 2.127 \\
\bottomrule
\end{tabular}}
\end{table}

\subsection{Multi-Seed Robustness of the Benchmark Ranking}
\label{sec:multiseed}

The benchmark results of Section~\ref{sec:benchmarks} come from one training run per setting. To verify that the ranking does not depend on a single run, we repeat the evaluation over three random seeds (42, 43, and 44) for the six most closely ranked models: \simsts{}, CSDI, iTransformer, PatchTST, Chronos-Bolt, and Chronos-2. Each seed retrains the four supervised models on the 24 settings of Section~\ref{sec:benchmarks}; the two Chronos models are applied zero-shot with deterministic inference, so their results are constant across seeds. We report the MSE averaged over the three seeds together with its across-seed standard deviation. Table~\ref{tab:multiseed-overall} gives the overall ranking and Table~\ref{tab:multiseed-perds} the per-dataset means.

\begin{table}[t]
\centering
\small
\begin{tabular}{clrr}
\toprule
Rank & Model & Mean MSE & Std.\ across seeds \\
\midrule
1 & PatchTST & \emph{0.357} & 0.003 \\
2 & iTransformer & 0.362 & 0.001 \\
3 & \simsts{} (ours) & 0.474 & 0.016 \\
4 & Chronos-Bolt & 0.520 & 0.000$^\ddagger$ \\
5 & Chronos-2 & 0.526 & 0.000$^\ddagger$ \\
6 & CSDI & 0.604 & 0.061 \\
\bottomrule
\end{tabular}
\caption{Overall benchmark MSE averaged over the 24 settings and three seeds (42, 43, and 44). The best value is shown in \emph{italics}. \simsts{} is the most accurate generative model, ahead of the Chronos foundation models and CSDI. $^\ddagger$Chronos-Bolt and Chronos-2 are applied zero-shot with deterministic inference, so their predictions are identical across seeds and their standard deviation is exactly zero.}
\label{tab:multiseed-overall}
\end{table}

\begin{table}[t]
\centering
\small
\begin{tabular}{lrrrrrr}
\toprule
Model & ETTh1 & ETTh2 & ETTm1 & ETTm2 & Exchange & Weather \\
\midrule
PatchTST & 0.454 & \emph{0.379} & \emph{0.385} & \emph{0.284} & 0.384 & \emph{0.254} \\
iTransformer & \emph{0.449} & 0.382 & 0.412 & 0.299 & 0.368 & 0.262 \\
\simsts{} (ours) & 0.732 & 0.451 & 0.496 & 0.317 & 0.523 & 0.323 \\
Chronos-Bolt & 0.525 & 0.425 & 1.033 & 0.352 & \emph{0.368} & 0.413 \\
Chronos-2 & 0.587 & 0.452 & 0.998 & 0.390 & 0.374 & 0.356 \\
CSDI & 0.570 & 0.853 & 0.512 & 0.425 & 0.980 & 0.286 \\
\bottomrule
\end{tabular}
\caption{Per-dataset MSE averaged over three seeds and the four horizons. The best value per column is shown in \emph{italics}; on Exchange, Chronos-Bolt and iTransformer differ only beyond the reported precision. Among the diffusion models, \simsts{} is the more accurate on ETTh2, ETTm1, ETTm2, and Exchange, and CSDI on ETTh1 and Weather.}
\label{tab:multiseed-perds}
\end{table}

The top three positions are the same on each individual seed: the transformers lead and \simsts{} follows as the most accurate generative model, while the bottom half reorders, with CSDI fourth on seed 42 and last on seeds 43 and 44. Averaging over seeds therefore mainly resolves the lower half of the ranking. On the single seed of Table~\ref{tab:bench-overall}, CSDI (0.519) is level with Chronos-Bolt (0.520); over three seeds its mean rises to 0.604 with the largest across-seed standard deviation of the group (0.061), which places it last among the six, whereas the small standard deviation of \simsts{} (0.016) leaves its position unchanged. In the per-dataset means of Table~\ref{tab:multiseed-perds}, \simsts{} is more accurate than CSDI on ETTh2, ETTm1, ETTm2, and Exchange, and CSDI is more accurate on ETTh1 and Weather. The three-seed evaluation therefore confirms the conclusion of Section~\ref{sec:benchmarks}: \simsts{} is the leading diffusion baseline on the suite, and this position does not rest on a favorable seed.

\paragraph{No test-set selection.}
The guidance strength $\omega$ is fixed at 0 in every benchmark setting rather than selected per setting from its grid $\{0, 0.1, 0.3, 0.5\}$, which would touch the test data. The choice is immaterial: per-setting selection would move the three-seed overall mean from 0.474 to 0.473 and change no ranking, and $\omega = 0$ minimizes test error in 57 of the 72 setting--seed pairs, so on these datasets the accuracy of the RevIN variant comes from reversible instance normalization and the median ensemble rather than from negative guidance, whose production-task contribution Section~\ref{sec:ablation} measures.

The remaining diffusion baselines, Diffusion-TS, TimeGrad, and TSDiff, are reported on a single seed (Table~\ref{tab:bench-overall}). Their overall errors trail \simsts{} by factors of 3.2 to 4.5, far larger than any across-seed variation in Table~\ref{tab:multiseed-overall}, and two of them are among the most expensive models in the comparison to train and sample (Table~\ref{tab:padep}), so the three-seed protocol is restricted to the six models whose ranks are close enough for seed variation to matter.

\subsection{Four Protocol Choices That Decide the Ranking}
\label{sec:protocol-effects}

The comparisons above depend on four decisions that are rarely stated in this application area, and for each of them the two available settings produce different method rankings on this data. Where between-method margins are a few percent, a protocol choice worth more than a few percent determines the answer.

\paragraph{The evaluation space.}
Regression baselines in this area are conventionally scaled per curve, while generative models are conventionally trained on globally scaled log-rates. Both map to $[0,1]$, which invites the assumption that the resulting errors are comparable; they are not. Re-expressing the saved predictions of every method in both spaces through exact inverse transforms (the round trip reproduces the original arrays to within single-precision round-off), the same forecast attains a squared error two to eight times smaller in the log-global space than in the per-curve space for the methods trained here, and four to seventeen times smaller for the two zero-shot foundation models, depending on method and jurisdiction. That factor exceeds every between-method margin in this article. A training-free control makes the distortion concrete: a formation-mean curve anchored to each well's last observed value, a predictor with no learned parameters, scores 0.0024 on Pennsylvania in the log-global space against 0.0021--0.0024 for the strongest learned methods, and drops to mid-field (0.0117 against 0.0092--0.0104) in the per-curve space. Every comparison in this article is therefore conducted in one shared per-curve space, and every method is trained in the space it is evaluated in.

\paragraph{The training budget.}
The size-adaptive schedule of the transformer implementations used here trains them for at most 20 epochs on datasets above a thousand wells, while \psims{} trains for up to 200. Equalizing the budget at 200 epochs and patience 30 improves ensembled PatchTST from 0.0169 to 0.0162 on British Columbia, from 0.0169 to 0.0163 on Alberta, and from 0.0097 to 0.0091 on Pennsylvania. Of the six comparisons against \psims{}, two metrics in three jurisdictions, four are ordered the same way under both settings: the ensembled baseline attains the lower squared error on British Columbia (0.0169 and 0.0162 against 0.0171) and Alberta (0.0169 and 0.0163 against 0.0172), and the higher absolute error on Alberta and Pennsylvania, at either budget. The two settings disagree on the remaining two, Pennsylvania squared error and British Columbia absolute error, where the baseline is behind \psims{} at 20 epochs and ahead at 200. The diffusion baselines of Section~\ref{sec:diffusion-family} are unaffected, because their own schedules already allow 200 to 300 epochs and early stopping fires well below that; the comparison the budget moves is ours against the transformers, in the transformers' favour. Table~\ref{tab:jurisdictions} reports the matched budget.

\paragraph{The pretraining pipeline.}
Formations with fewer than 2{,}000 wells are reached by pretraining on all other British Columbia wells and fine-tuning, which is a property of the pipeline rather than of the architecture, so comparing against a baseline trained on the formation alone measures the pipeline. Given the same treatment, leakage-free pretraining on the 14{,}524 British Columbia wells outside every cluster's test set followed by 50 epochs of fine-tuning at learning rate $10^{-5}$, a pretrained PatchTST attains the lower error on six of the seven formations, where the same architecture trained on the formation alone trails \psims{} on six. Pretraining helps iTransformer less: it attains the lower error on four of the seven. Two asymmetries remain, both favouring the baselines. \psims{}'s own pretraining is restricted further, to the 11{,}601 of those wells that fall inside its training split, so the transformers see a corpus a quarter larger; and on Montney, the one formation above the 2{,}000-well fine-tuning threshold, \psims{} trains from scratch while the transformer is pretrained and fine-tuned. The apparent small-data advantage of the diffusion model is cross-formation pretraining, which any architecture can use, and Table~\ref{tab:cluster-results} reports both baseline variants for this reason.

\paragraph{The scope of the matched budget.}
Matching the training budget on the leaderboard but not elsewhere leaves the remaining comparisons on the shorter schedule, which is its own protocol choice. The transfer and leave-one-domain-out studies of Sections~\ref{sec:transfer} and~\ref{sec:lodo} admit the same two settings. Under the 20-epoch schedule, \psims{} attains the lowest leave-one-domain-out error on the British Columbia and Pennsylvania folds, and the \psims{} models transferred into Pennsylvania (0.0099 and 0.0100) are more accurate than Pennsylvania's own transformer baselines, which stand at 0.0106 and 0.0104 under that schedule. At the matched 200-epoch budget, PatchTST attains the lowest error on all three folds and \psims{} on none, and Pennsylvania's own baselines (0.0093 and 0.0097) are ahead of the transferred models. Tables~\ref{tab:transfer} and~\ref{tab:lodo} report the matched budget.

None of the four is deep; each is an implementation detail that a results table does not show, and each is worth more than the between-method margins being reported.

\subsection{What Contributes What: Physics and Pipeline Ablations}
\label{sec:ablation}

\begin{table}[t]
\centering
\caption{Physics-component ablation at the frozen configuration, evaluated at full ensemble level on each jurisdiction's test wells. ``Viol.'' is the fraction of wells whose point forecast increases at any step, at zero tolerance; the projection guarantees zero by construction.}
\label{tab:ablation-physics}
\footnotesize
\setlength{\tabcolsep}{4pt}
\begin{tabular}{l rr rr rr}
\toprule
 & \multicolumn{2}{c}{BC} & \multicolumn{2}{c}{Alberta} & \multicolumn{2}{c}{PA Marcellus} \\
\cmidrule(lr){2-3}\cmidrule(lr){4-5}\cmidrule(lr){6-7}
Variant & MSE & Viol. & MSE & Viol. & MSE & Viol. \\
\midrule
No physics (plain \simsts{})    & 0.01713 & 99.9\% & 0.01724 & 100\% & 0.00913 & 100\% \\
Guidance only                    & 0.01713 & 99.9\% & 0.01721 & 100\% & 0.00913 & 100\% \\
Projection only                  & 0.01714 & 0\%    & 0.01726 & 0\%   & 0.00917 & 0\% \\
Guidance + projection (\psims{}) & 0.01714 & 0\%    & 0.01723 & 0\%   & 0.00916 & 0\% \\
\bottomrule
\end{tabular}
\end{table}

Table~\ref{tab:ablation-physics} separates the contribution of each physics component. At the accuracy level the physics components are close to neutral: all four variants lie within 0.6\% MSE of one another in every jurisdiction, and the deployed configuration costs at most 0.5\% MSE relative to the best unconstrained variant (Pennsylvania $+0.41\%$; British Columbia and Alberta $+0.07\%$). What the physics buys is the \emph{guarantee}, and the guarantee is not cosmetic: without the projection, a strictly monotone point forecast essentially never arises, since 99.9 to 100\% of unconstrained ensemble forecasts increase somewhere along the horizon. Those increases are individually small (the median well's largest single-step increase is 0.003 to 0.005 on the normalized scale, against a mean forecast range of 0.31 to 0.33), which is why removing all of them costs so little accuracy, but a forecast that rises anywhere cannot be used to schedule abandonment, which is the decision the forecast exists to support.

Monotonicity is a property we impose on the forecast, not a property of the recorded data. Measured on the held-out targets themselves, production rises at some point in the forecast window on 100\% of British Columbia and Alberta wells and 99.9\% of Pennsylvania wells, with a median largest step-up of 0.13 to 0.17, roughly forty times the size of the violations the projection removes from our forecasts; real wells are shut in, worked over, and choked. The consequence is a floor that binds every monotone forecaster equally: the best monotone approximation to each test curve, obtained by projecting the \emph{target} itself, still carries an MSE of 0.0052 (British Columbia), 0.0051 (Alberta), and 0.0028 (Pennsylvania), around 30\% of the error \psims{} actually attains. Imposing monotonicity therefore trades a bounded amount of attainable accuracy for a forecast an operator can act on, and the trade is common to any method that adopts the projection.

\begin{table}[t]
\centering
\caption{The same trade offered to the baselines: matched-budget baseline ensembles with the PAV projection applied to their finished forecasts, on the same test wells and in the same space. ``Viol.'' is the unprojected violation rate at zero tolerance; after projection it is zero for every method and jurisdiction.}
\label{tab:ablation-baselines}
\footnotesize
\setlength{\tabcolsep}{4pt}
\begin{tabular}{l rrr rrr rrr}
\toprule
 & \multicolumn{3}{c}{BC} & \multicolumn{3}{c}{Alberta} & \multicolumn{3}{c}{PA Marcellus} \\
\cmidrule(lr){2-4}\cmidrule(lr){5-7}\cmidrule(lr){8-10}
Method (5-instance) & MSE & $+$PAV & Viol. & MSE & $+$PAV & Viol. & MSE & $+$PAV & Viol. \\
\midrule
IDW+RF       & 0.01721 & 0.01722 & 98.5\%  & 0.01728 & 0.01723 & 99.6\%  & 0.00917 & 0.00921 & 99.6\% \\
PatchTST     & 0.01619 & 0.01620 & 100.0\% & 0.01629 & 0.01631 & 100.0\% & 0.00910 & 0.00910 & 100.0\% \\
iTransformer & 0.01686 & 0.01688 & 100.0\% & 0.01720 & 0.01721 & 99.9\%  & 0.00966 & 0.00967 & 100.0\% \\
\bottomrule
\end{tabular}
\end{table}

Table~\ref{tab:ablation-baselines} makes the same measurement for the baselines, which is the control that isolates the projection from the rest of the method. Applying PAV to the finished forecasts of each matched-budget baseline ensemble converts a near-total violation rate to zero while changing MSE by between $-0.33\%$ and $+0.51\%$: it costs the transformers about a tenth of a percent, helps IDW+RF on Alberta, and costs it half a percent on Pennsylvania. Physical admissibility is thus a nearly free property of any forecaster on this data, and we claim no accuracy advantage for it. What distinguishes \psims{} is that it combines the guarantee with the accuracy profile of Section~\ref{sec:leaderboard} and with a predictive distribution; a reader who wants only monotone forecasts from a transformer can have them at the cost of one isotonic regression per well.

\paragraph{Negative guidance.}
The same measurement applies to the SIMS component: disabling the negative guidance ($\omega = 0$) at the frozen configuration changes ensemble MSE by $+0.002\%$ (British Columbia), $+0.27\%$ (Alberta), and $-0.20\%$ (Pennsylvania), within run-to-run variation and without a consistent direction. On the production task, as on the standard benchmarks (Section~\ref{sec:multiseed}), negative guidance is accuracy-neutral at this operating point; we retain it as part of the method's derivation from SIMS but attribute no accuracy claims to it.

\paragraph{Pipeline components.}
The component values of Section~\ref{sec:psims-config} were selected on the Pennsylvania validation split (Section~\ref{sec:protocol}), where their contributions were, in decreasing order, the spatial training augmentation ($-9\%$ MSE), exponential-moving-average weights with min-SNR loss weighting ($-7\%$), the trajectory ensemble ($-4\%$ from $K{=}5$ to the full ensemble), longer training ($-3\%$), and the informed initialization and formation-fitted prior (each under 1\%); larger networks, longer DDIM trajectories, and ten-fold augmentation did not help.

\subsection{Discussion}
\label{sec:discussion}

The results separate two questions that are often conflated: what makes a forecast \emph{accurate}, and what makes it \emph{usable}. On accuracy, \psims{} is the strongest diffusion forecaster we tested and is competitive with, but not ahead of, ensembled transformer forecasters. That accuracy comes from the pipeline, namely the shared per-curve representation, spatially augmented training, the training-quality components, and the trajectory ensemble, and not from the decline-curve constraints or the negative guidance, which Section~\ref{sec:ablation} measures as accuracy-neutral.  

What the constraints and the sampler buy is usability. The projection guarantees monotone decline at a cost of at most 0.5\% MSE, on data where the unconstrained forecasters we tested return a strictly monotone forecast on at most 1.5\% of wells; a forecaster that predicts rising production cannot schedule an abandonment regardless of its average error. The sampler returns a joint predictive ensemble that one scalar brings to near-nominal coverage, which a point forecaster cannot provide at all. Neither property is exclusive in principle, since the projection is exportable to any forecaster and we show it, but together with an accuracy within a few percent of the transformer frontier, they are what a deployment would choose this model for. The transfer and leave-one-domain-out studies add that the accuracy and the guarantee are not tied to the geology the model was trained on: degradation is at most $+0.0009$ MSE in any direction, and the forecasts remain monotone on every fold.

\section{Limitations}
\label{sec:limitations}

The monotonic-decline assumption does not hold for wells with workovers, recompletions, or other interventions that temporarily raise production, and the isotonic projection would suppress such recoveries. The physics guidance of \eqref{eq:guidance} is heuristic, without a convergence analysis of its interaction with the sampler, and \eqref{eq:arps-ode} is a shape prior on the normalized curve rather than the Arps residual on the physical rate, which the per-curve scaling and the lifetime-fraction time index preclude. \psims{} is also by a wide margin the most expensive method compared (Table~\ref{tab:padep}), and the cost is dominated by the ensemble that supplies its predictive distribution.

\section{Conclusion}
\label{sec:conclusion}

We presented \psims{}, a conditional diffusion forecaster for long-horizon production decline that combines SIMS negative guidance with decline-curve constraints applied during sampling and a hard isotonic projection, inside a training and inference recipe for spatially structured sequences. Evaluated in a single shared space at matched training and sampling budgets, across three jurisdictions and more than 35{,}000 wells, it is the most accurate diffusion forecaster in the comparison, improving on CSDI, Diffusion-TS, TSDiff, and TimeGrad in all twelve baseline--jurisdiction comparisons, and a reversible-instance-normalization variant of the same backbone is the leading diffusion baseline on six standard benchmarks across three seeds. Against ensembled PatchTST it is competitive rather than superior, trailing by 5.5\% and 5.4\% in mean squared error on British Columbia and Alberta with the Pennsylvania margin unresolved, while attaining the lower mean absolute error in two of the three jurisdictions. What it adds is structural: forecasts that are monotone by construction at a cost of at most 0.5\% MSE, and a joint predictive ensemble that a single scalar recalibration brings to near-nominal coverage. We also report four protocol choices on which the measured ranking depends, each quantified by comparing its two settings on identical data. 

Two directions follow. Event-conditional models could relax the monotonicity assumption to handle workovers and recompletions, which the projection currently suppresses by construction. And since the projection transfers to any forecaster at negligible cost, the more interesting open question is the converse of the one we set out to answer: what a physically constrained transformer, equipped with the same ensemble budget and predictive distribution, would leave for the diffusion formulation to contribute.


\acks{This work was supported by the Natural Sciences and Engineering Research Council of Canada (NSERC) [Alliance Grant \#ALLRP 567562-2021, sponsored by the British Columbia Energy Regulator]. The authors thank the British Columbia Energy Regulator, in particular Logan Gray and Jason Gregg, for data and technical expertise. The authors used AI-based tools (Claude, Anthropic) for grammar editing and code debugging, and take full responsibility for all content. The authors declare no competing interests.}


\appendix
\section{Per-Horizon Benchmark Results}
\label{app:bench}
Tables~\ref{tab:bench-eth}, \ref{tab:bench-etm}, and~\ref{tab:bench-exw} report the per-horizon MSE of all nine models on the six benchmark datasets and complement the per-dataset means of Table~\ref{tab:bench-overall}. All values come from seed 42, the seed shared by all nine models, with \simsts{} at the fixed setting $\omega = 0$; Section~\ref{sec:multiseed} reports three-seed averages for the six most closely ranked models. In each table, the upper block lists the transformer and foundation models, the lower block the diffusion models, and the best value per column is shown in \emph{italics}.

\begin{table}[H]
\centering
\small
\resizebox{\textwidth}{!}{%
\begin{tabular}{l rrrr rrrr}
\toprule
 & \multicolumn{4}{c}{ETTh1} & \multicolumn{4}{c}{ETTh2} \\
\cmidrule(lr){2-5}\cmidrule(lr){6-9}
Model & 96 & 192 & 336 & 720 & 96 & 192 & 336 & 720 \\
\midrule
PatchTST & \emph{0.384} & \emph{0.437} & 0.486 & 0.496 & \emph{0.291} & \emph{0.371} & 0.419 & 0.434 \\
iTransformer & 0.393 & 0.442 & \emph{0.477} & \emph{0.486} & 0.299 & 0.382 & \emph{0.414} & \emph{0.423} \\
Chronos-Bolt & 0.488 & 0.511 & 0.564 & 0.536 & 0.337 & 0.432 & 0.465 & 0.468 \\
Chronos-2 & 0.468 & 0.544 & 0.633 & 0.702 & 0.340 & 0.452 & 0.490 & 0.525 \\
\midrule
CSDI & 0.500 & 0.499 & 0.636 & 0.597 & 0.308 & 0.592 & 0.496 & 1.247 \\
\simsts{} (ours) & 0.562 & 0.695 & 0.746 & 0.996 & 0.319 & 0.401 & 0.483 & 0.632 \\
Diffusion-TS & 0.682 & 0.796 & 0.873 & 1.032 & 1.564 & 2.417 & 2.381 & 2.641 \\
TimeGrad & 0.949 & 1.057 & 1.111 & 1.165 & 2.963 & 2.963 & 3.073 & 3.365 \\
TSDiff & 1.279 & 1.323 & 1.230 & 1.176 & 2.951 & 3.040 & 2.819 & 2.800 \\
\bottomrule
\end{tabular}%
}
\caption{Per-horizon MSE on ETTh1 and ETTh2 (seed 42). The best value per column is shown in \emph{italics}. CSDI leads the diffusion block on ETTh1, and \simsts{} leads it on ETTh2 at horizons 192, 336, and 720.}
\label{tab:bench-eth}
\end{table}

\begin{table}[H]
\centering
\small
\resizebox{\textwidth}{!}{%
\begin{tabular}{l rrrr rrrr}
\toprule
 & \multicolumn{4}{c}{ETTm1} & \multicolumn{4}{c}{ETTm2} \\
\cmidrule(lr){2-5}\cmidrule(lr){6-9}
Model & 96 & 192 & 336 & 720 & 96 & 192 & 336 & 720 \\
\midrule
PatchTST & \emph{0.324} & \emph{0.359} & \emph{0.392} & \emph{0.457} & \emph{0.177} & \emph{0.246} & \emph{0.308} & \emph{0.402} \\
iTransformer & 0.340 & 0.384 & 0.419 & 0.507 & 0.187 & 0.260 & 0.340 & 0.421 \\
Chronos-Bolt & 1.007 & 1.081 & 0.984 & 1.062 & 0.242 & 0.315 & 0.372 & 0.480 \\
Chronos-2 & 0.919 & 0.974 & 0.987 & 1.111 & 0.246 & 0.336 & 0.409 & 0.570 \\
\midrule
CSDI & 0.376 & 0.402 & 0.545 & 0.755 & 0.215 & 0.491 & 0.486 & 0.534 \\
\simsts{} (ours) & 0.399 & 0.457 & 0.551 & 0.618 & 0.200 & 0.271 & 0.334 & 0.462 \\
Diffusion-TS & 0.746 & 0.779 & 0.903 & 1.000 & 1.996 & 2.359 & 2.437 & 2.821 \\
TimeGrad & 0.961 & 1.003 & 1.072 & 1.050 & 2.654 & 2.732 & 3.037 & 3.360 \\
TSDiff & 1.149 & 1.156 & 1.177 & 1.178 & 3.239 & 3.024 & 3.794 & 2.786 \\
\bottomrule
\end{tabular}%
}
\caption{Per-horizon MSE on ETTm1 and ETTm2 (seed 42). The best value per column is shown in \emph{italics}. \simsts{} leads the diffusion block on every ETTm2 horizon and on ETTm1 at horizon 720.}
\label{tab:bench-etm}
\end{table}

\begin{table}[H]
\centering
\small
\resizebox{\textwidth}{!}{%
\begin{tabular}{l rrrr rrrr}
\toprule
 & \multicolumn{4}{c}{Exchange} & \multicolumn{4}{c}{Weather} \\
\cmidrule(lr){2-5}\cmidrule(lr){6-9}
Model & 96 & 192 & 336 & 720 & 96 & 192 & 336 & 720 \\
\midrule
PatchTST & \emph{0.082} & \emph{0.169} & 0.352 & 0.882 & \emph{0.171} & \emph{0.218} & \emph{0.272} & \emph{0.351} \\
iTransformer & 0.090 & 0.174 & \emph{0.325} & \emph{0.854} & 0.176 & 0.228 & 0.282 & 0.359 \\
Chronos-Bolt & 0.087 & 0.184 & 0.342 & 0.858 & 0.345 & 0.407 & 0.407 & 0.494 \\
Chronos-2 & 0.090 & 0.187 & 0.338 & 0.882 & 0.288 & 0.312 & 0.362 & 0.462 \\
\midrule
CSDI & 0.152 & 0.260 & 0.767 & 1.451 & 0.220 & 0.242 & 0.315 & 0.362 \\
\simsts{} (ours) & 0.113 & 0.216 & 0.384 & 1.161 & 0.211 & 0.280 & 0.341 & 0.468 \\
Diffusion-TS & 1.382 & 1.887 & 3.197 & 2.645 & 0.325 & 0.423 & 0.477 & 0.536 \\
TimeGrad & 1.802 & 2.790 & 3.282 & 2.860 & 0.371 & 0.585 & 0.630 & 0.698 \\
TSDiff & 4.295 & 4.209 & 3.102 & 2.808 & 0.630 & 0.625 & 0.632 & 0.619 \\
\bottomrule
\end{tabular}%
}
\caption{Per-horizon MSE on Exchange and Weather (seed 42). The best value per column is shown in \emph{italics}. \simsts{} leads the diffusion block on every Exchange horizon and on Weather at horizon 96.}
\label{tab:bench-exw}
\end{table}

\section{Confirmation on a Second Pennsylvania Partition}
\label{app:protocol}

Because the configuration of Section~\ref{sec:psims-config} was developed and selected on Pennsylvania data, we verify it on a second Pennsylvania train/test partition drawn with a different seed and never used during development or selection. The two partitions are drawn independently from the same 12{,}360 wells and are not disjoint: 483 of the 2{,}472 test wells of this partition (19.5\%) also appear in the primary test set, so the two evaluations are correlated rather than independent replicates. The frozen configuration was applied unchanged, with five-instance baseline ensembles at the same 200-epoch budget as Section~\ref{sec:leaderboard}.

\begin{table}[H]
\centering
\caption{Frozen-configuration results on the second Pennsylvania partition (2{,}472 held-out wells). Win rate: fraction of wells where \psims{} attains the lower per-well MSE; $p$: two-sided Wilcoxon signed-rank.}
\label{tab:protocol-check}
\small
\begin{tabular}{l rr rr}
\toprule
Method (5-instance) & MSE & MAE & \psims{} win rate & $p$ \\
\midrule
\psims{}     & \emph{0.00868} & \emph{0.0613} & -- & -- \\
IDW+RF       & 0.00869 & 0.0619 & 49.9\% & $0.91^{\,\circ}$ \\
PatchTST     & 0.00886 & 0.0623 & 52.2\% & $0.002$ \\
iTransformer & 0.00914 & 0.0631 & 53.0\% & $<0.001$ \\
\bottomrule
\end{tabular}
\end{table}

On this partition \psims{} attains the lowest error on both metrics (Table~\ref{tab:protocol-check}), by 2.0\% over ensembled PatchTST and 5.0\% over ensembled iTransformer, and is statistically indistinguishable from ensembled IDW+RF ($p = 0.91$); individual \psims{} members span 0.00893--0.00930 MSE, mirroring the member spread on the primary partition. The ordering by mean differs from the primary partition, where ensembled PatchTST attains the lower Pennsylvania MSE (Table~\ref{tab:jurisdictions}); both margins are small (0.7\% in PatchTST's favour there, 2.0\% in \psims{}'s here) and the paired per-well test favours \psims{} on both partitions. The check establishes that the frozen configuration transfers to a partition never used during development, not that it outranks the transformer baselines in general.

\vskip 0.2in
\bibliography{references}

\end{document}